\pdfoutput=1
\documentclass[11pt]{article}

\usepackage[preprint]{acl}
\usepackage{times}
\usepackage{latexsym}

\usepackage[T1]{fontenc}
\usepackage[utf8]{inputenc}

\usepackage{microtype}

\usepackage{inconsolata}

\usepackage{graphicx}
\usepackage{subfigure}
\usepackage{booktabs} % for professional tables
\usepackage{outlines}
\usepackage{subcaption}
\usepackage{wrapfig}
\usepackage{enumitem}

\usepackage[table]{xcolor}

\definecolor{lightblue}{RGB}{200, 230, 255} % Soft pastel blue
\definecolor{lightred}{RGB}{255, 210, 210}

\usepackage{adjustbox}
\usepackage{booktabs}

\usepackage{hyperref}

\usepackage{amsmath}
\usepackage{amssymb}
\usepackage{mathtools}
\usepackage{amsthm}
\usepackage{multirow}
\usepackage{makecell}
\usepackage{tikz}
\usepackage{adjustbox}
\usepackage[dvipsnames]{xcolor}

\usepackage[capitalize,noabbrev]{cleveref}

\theoremstyle{plain}

\theoremstyle{definition}

\theoremstyle{remark}

\usepackage[textsize=tiny]{todonotes}

\title{Mechanisms of Prompt-Induced Hallucination in Vision–Language Models}

\author{
    William Rudman\thanks{Equal contribution. Order determined by coin flip.}$^{1}$, Michal Golovanevsky$^{* 2}$, Dana Arad$^{3}$, \\
    {\bf Yonatan Belinkov$^{3, 5}$},
    {\bf Ritambhara Singh$^{2}$},
    {\bf Carsten Eickhoff$^{4}$}, {\bf Kyle Mahowald$^{1}$}\\
    $^{1}$The University of Texas at Austin, $^{2}$Brown University, $^{3}$Technion, \\
    $^{4}$University of Tübingen, $^{5}$Harvard University\\
    \texttt{michal\_golovanevsky@brown.edu} \& \texttt{william.rudman@utexas.edu} \\
}

\begin{document}
\maketitle
\begin{abstract}
\renewcommand{\thefootnote}{\fnsymbol{footnote}}
Large vision–language models (VLMs) are highly capable, yet often hallucinate by favoring textual prompts over visual evidence. We study this failure mode in a controlled object-counting setting, where the prompt overstates the number of objects in the image (e.g., asking a model to describe \textit{four} waterlilies when only \textit{three} are present).
At low object counts, models often correct the overestimation, but as the number of objects increases, they increasingly conform to the prompt regardless of the discrepancy.
Through mechanistic analysis of three VLMs, we identify a small set of attention heads whose ablation substantially reduces prompt-induced hallucinations (PIH) by at least 40\% without additional training.
Across models, PIH-heads mediate prompt copying in model-specific ways. We characterize these differences and show that PIH ablation increases correction toward visual evidence.
Our findings offer insights into the internal mechanisms driving prompt-induced hallucinations, revealing model-specific differences in how these behaviors are implemented. \footnote{Code available at: \url{https://github.com/michalg04/prompt-induced_hallucinations.git}}

\end{abstract}

\section{Introduction}

\begin{figure}[h!]
    \centering
    \includegraphics[width=0.9\columnwidth]{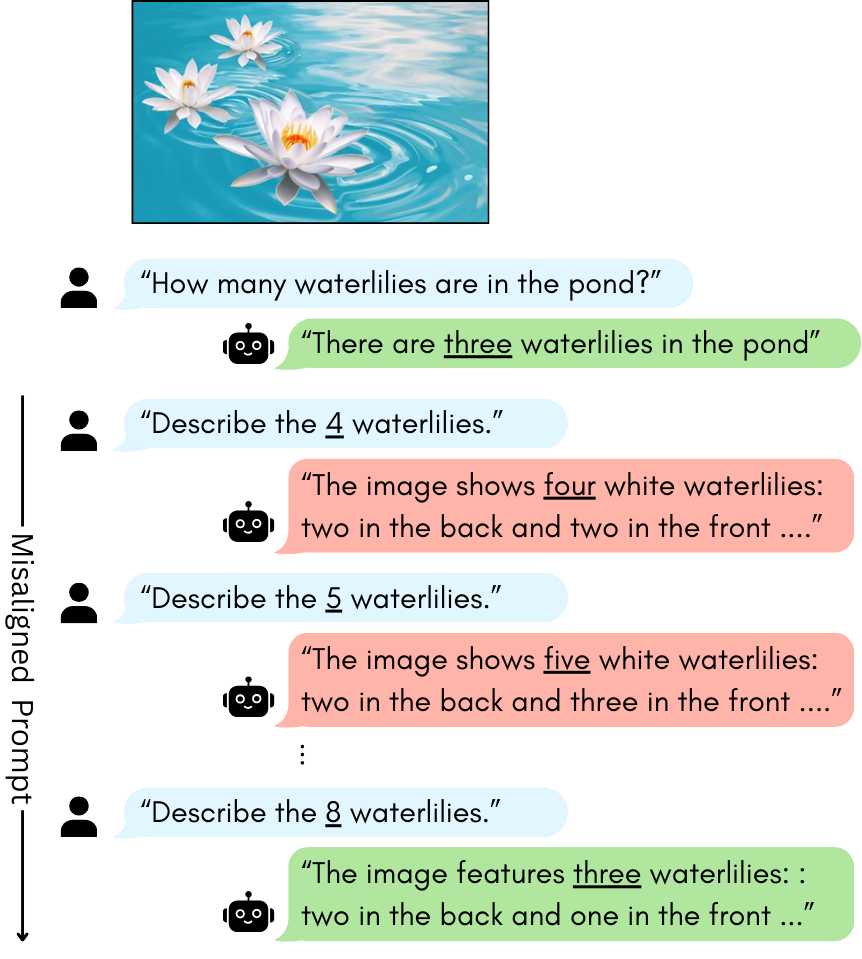}
    \caption{\textbf{Example of prompt-induced hallucination (PIH) in numerical reasoning.} The model correctly answers the baseline counting question. However, it conforms to the prompt and hallucinates additional waterlilies when asked to describe a mismatched number of objects. As the mismatch between the prompt and visual evidence grows, the model increasingly favors the image and recovers the correct count.}
    \label{fig:main}
\end{figure}

Vision–language models (VLMs) often follow textual prompts even when they contradict visual evidence \cite{deng2025words, zhang2025modalities, hua2025vision}. These hallucinations are a growing concern as VLMs are deployed in real-world settings with noisy or inaccurate user inputs. To better understand this behavior, we study \textit{prompt-induced hallucinations} (PIH) in a controlled object-counting setting, where prompts systematically overestimate the number of objects in an image.

% one sentence on how these are hallucinations and not just prompt following. 
% Bar bars in fig fatter

Figure~\ref{fig:main} illustrates the phenomenon where, although the model correctly counts three waterlilies when asked directly, it hallucinates additional, nonexistent flowers when asked to describe a mismatched count (e.g., ``Describe the four waterlilies'').
This undesirable behavior is unlikely to be driven by limitations in numerical reasoning \cite{zhang2023number,rahmanzadehgervi2024vision, rudman2025forgotten}, since the model correctly solves the underlying counting task. Instead, this is consistent with a known tendency of VLMs to rely on textual prompts over conflicting visual evidence \cite{deng2025words}. 

To characterize PIH, we study \textbf{when} such errors occur and \textbf{how} they can be reduced via targeted analysis of attention-heads.
Across models, PIH outputs follow a consistent and structured pattern.
When the ground-truth number of objects exceeds four, VLMs are more likely to conform to the prompt and hallucinate missing objects.
In contrast, at lower object counts, models are more likely to copy the prompt under small discrepancies but transition to visually grounded responses as the mismatch increases (\cref{fig:main}).

Through mechanistic analysis, we identify attention heads associated with PIH.
Ablating these heads reduces hallucinations and restores visually grounded counts by up to \textit{$54$}\%.
Importantly, this intervention does not require additional data or training and alters only the object count, leaving the described objects unchanged. To test generalization beyond counting, we apply the same PIH head ablations to a color identification task \cite{pvp} in which the prompt specifies an incorrect color.
We find that these heads similarly mediate PIH in the color domain, mitigating prompt-color copying by up to $94.25$\%.

We further investigate the functional role of PIH attention heads and find that their ablation consistently reduces prompt-based copying, increases correction toward ground-truth responses, and shifts attention toward the image. While these effects are shared across all tested models, the mechanisms by which copying is suppressed differ.
%We draw this distinction by analyzing response formatting (exact, format, and soft copying), attention shifts between text and image tokens, and underlying probabilities, which together reveal model-specific behavior. Across models, analysis of response probabilities in the absence of prompt conflict shows that PIH ablation consistently reduces exact prompt-copying. Depending on the model, this reduction manifests as increased format or soft copying, but in all cases, it is accompanied by greater reliance on image tokens and improved correctness.
We make this distinction by examining how models format their responses, how attention shifts between text and image inputs, and how output probabilities change after ablation. Across models, these analyses show that PIH ablation consistently reduces direct copying from the prompt. While the way this reduction appears varies by model, it is always accompanied by increased reliance on visual information and improved correctness.
LLaVA-OneVision shows the strongest reduction in prompt-following after PIH-heads ablation, with clear suppression of prompt-consistent formatting, together with the largest increase in image attention and improved general counting performance. 

Our contributions are threefold:
\begin{enumerate}[topsep=0pt, itemsep=0ex]
    \item We introduce and analyze PIH, a failure mode in which VLMs prioritize prompt information over conflicting visual evidence. 
    % We show that PIH is most pronounced when confidence in visual grounding is low 
    (\cref{sec:pih}).
    \item We identify a small subset of attention heads whose ablation substantially reduces PIH across models and tasks, without additional training or performance degradation on aligned prompts (\cref{sec:mechanisms} \& \cref{sec:pih_generalization}).
    \item We characterize the function of PIH-heads and show their ablation suppresses prompt copying while increasing correction toward visual evidence. (\cref{sec:pih_functions}).
\end{enumerate}

\section{Related Work}
\subsection{Prompt-Induced Hallucinations in LLMs}
A substantial body of work has investigated hallucinations in large language models (LLMs) \cite{ji2023survey, tonmoy2024comprehensive, huang2025survey}. 
Prior work has shown that hallucinations can be reliably induced through out-of-distribution prompting \cite{yao2023llm}, long context windows \cite{li2024measuring, xu2024earth}, or the use of explicit personas \cite{joshi2024personas, simhi2025hack}. 
However, these settings often reflect adversarial or non-realistic usage patterns, rather than the models' behavior under typical user interactions.  
Previous studies have demonstrated that LLMs may produce incorrect outputs when prompts contradict internal knowledge \cite{xie2023adaptive, xu2024knowledge} or contain errors subsequently repeated or amplified by the model \cite{zhang2023language, simhi2025hack}. 
This behavior has been linked to cognitive-like biases, including anchoring bias, where models rely disproportionately on information introduced in the prompt \cite{jones2022capturing, echterhoff2024cognitive, malberg2025comprehensive}, and sycophancy, where models conform to user assumptions despite conflicting evidence \cite{sharma2024towards}.

% Hallucinations are often divided into contextual, i.e., triggered by properties of the input \cite{}, and parametric hallucinations, stemming from the models' internal knowledge \cite{}.
% We characterize PIH as a type of contextual (also known as close-book) hallucinations. 

\begin{figure*}[h]
    \centering
    \includegraphics[width=0.85\textwidth]{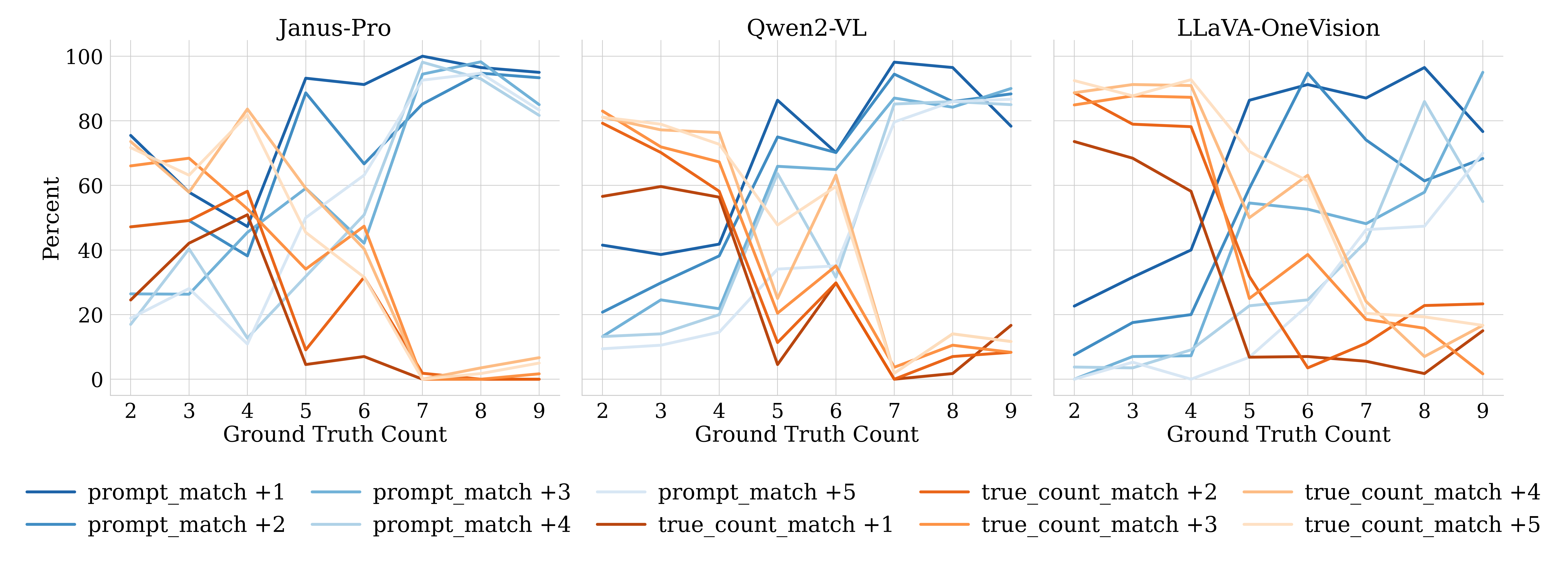}
    \caption{\textbf{PIH rates across different numbers of ground-truth objects.} Blue lines show the percentage of outputs matching the prompted number for different offsets, $k$. Orange lines show matches to the true count. As the object count increases, models stop correcting the prompt.}
    \label{fig:confirmation}
\end{figure*}

\subsection{Modality Conflicts in VLMs}
Vision–language models (VLMs) integrate textual and visual information, introducing additional complexity.
Prior work shows that VLMs achieve higher accuracy on textual tasks compared to visual equivalents \cite{frank2021vision, fu2024isobench, van2025same}. This gap is attributed to distinct internal mechanisms for visual and textual processing \cite{nikankin2025same, notice} and to modality-specific parametric knowledge \cite{zhu2024unraveling}.
VLMs also struggle to disentangle different signals, both across their input modalities \cite{hua2025vision} and between their input and parametric knowledge \cite{DBLP:conf/emnlp/TangYZY23, pvp, ortu2025seeing}.

Recent work investigated how VLMs perform under inconsistent or conflicting textual and visual inputs \cite{yan2025multimodal, zhao2025color}.
Given conflicting information, current VLMs tend to disproportionately prefer textual information \cite{salin2022vision, deng2025words}, which can manifest in sycophantic behavior \cite{pi2025pointing}. 
Further exploring modality conflicts, \citeauthor{zhang2025modalities} investigated modality preference through the lens of unimodal uncertainty.
In contrast, our work studies modality conflicts stemming from minor, realistic inconsistencies and examines a more complex task that more closely reflects real-world usage.

% \subsection{Specialized Attention Mechanisms}
% A growing line of work aims to identify task-specific model components \cite{hanna2023does, nikankin2024arithmetic}, such as attention heads, that play specialized functional roles \cite{zheng2024attention}.  
% In LLMs, studies have found induction heads to play a critical role in copying information from the prompt \cite{olsson2022context, wang2022interpretability, musat2025emergence} and have identified different heads used for verbatim copying vs. concept copying \cite{feucht2025dual}.
% In VLMs, studies have identified specialized attention heads, including heads associated with safety-related behaviors \cite{zheng2025spot}, supporting specific tasks such as OCR \cite{baek2025large}, or critical for disentangling visual and textual information \cite{hua2025vision}.

\section{Prompt-Induced Hallucinations in Object Counting}
\label{sec:pih}
In this section, we present Prompt-Induced Hallucinations (PIH) and study when they manifest in the context of object counting. Object counting provides a clean setting for isolating prompt–image conflicts: ground-truth answers are verifiable, numerical deviations can be precisely controlled, and model errors can be explicitly measured. While a language model could legitimately answer the question "Describe 4 waterlilies" as part of an instruction-following task without hallucinating, for VLMs these responses constitute hallucinations: (1) we explicitly ground prompts with the phrase ``in the image,'' and (2) the models describe additional objects not present in the image (see examples in Table~\ref{tab:app_qualitative_knockout_examples}).

\subsection{Experimental Setup}
We investigate three VLMs covering a range of state-of-the-art model families: LLaVA-OneVision-7B~\cite{llava-onevision}, Qwen2-VL-7B~\cite{qwen2-vl}, and Janus-Pro-7B~\cite{janus-pro}.
% , which have shown competitive counting performance \cite{rudman2025forgotten}.
% These models were selected to cover a range of current state-of-the-art multimodal architectures, including established families like LLaVA, and emerging MLLMs like QwenVL and Janus Pro.
We study counting using \textit{CountBench}~\cite{paiss2023teaching}, a benchmark designed to measure object-count understanding in VLMs.
For each image with ground-truth object count $N$, we first present a \emph{baseline prompt}, $P_{B} = $ ``How many [objects] are there in the image?'', retaining only samples where the model correctly predicts $N$.
We then use \emph{misaligned prompts} asking the model to ``Describe the $N+k$ [objects] in the image,'', where $k \in \{1,2,\ldots,5\}$ denotes the \emph{discrepancy distance} between the prompt and the visual input.
% We allow the model to generate long-form responses up to $75$ tokens and use a rule-based algorithm to determine accuracy by extracting the first non-negated numerical reference in the response (e.g., excluding constructions such as ``there are not eight waterlilies'').
PIH occur when the model describes $N+k$ objects without correcting the count, despite contradictory visual evidence. 
See \cref{app:exp_setup} for additional details.

\subsection{PIH at Different Object Counts}
Figure~\ref{fig:confirmation} shows the proportion of outputs that either match the prompted count ($N+k$, blue curves) or the true count ($N$, orange curves) across a range of object quantities and prompt offsets ($k$).
The $x$-axis denotes the true number of objects ($N$), and the $y$-axis shows the fraction of responses aligned with either the prompt or ground truth.
We observe a consistent pattern across all three models.
For images containing small numbers of objects ($2$-$4$ objects), models typically resist misleading prompts and report the true count, demonstrated by the orange curves being mostly higher than the blue curves. 
However, models still produce PIH in approximately $20$–$40$\% of cases, which primarily occur when the discrepancy is small ($N+1$). 
% For example, given an image containing two dogs, models often accept a prompt describing three dogs, yet reject larger discrepancies and explicitly correct the count to ground truth.

Beyond four objects, this behavior changes.
Models increasingly conform to the prompt regardless of the discrepancy size, with prompt-matching responses approaching $80$–$90$\% even when the prompt substantially overestimates the true count. Accordingly, ground truth responses drop to near zero.
In these cases, the discrepancy distance ceases to play a critical role, as shown by the convergence of the blue prompt-aligned lines.
%For images containing a larger number of objects, models tend to copy the number specified in the misaligned prompt regardless of the magnitude of the discrepancy. 
% For example, given an image with nine cats, the model may assert that there are ten, eleven, or even fourteen cats, when prompted to describe the $9+k$ cats.
This effect persists for extreme offsets (with $k \in \{10, 20, 50\}$), with models readily describing, for example, fifty-nine cats in an image containing only nine (more details in \cref{app:large_offsets}).

\begin{figure}
    \centering
    \includegraphics[width=0.95\linewidth]{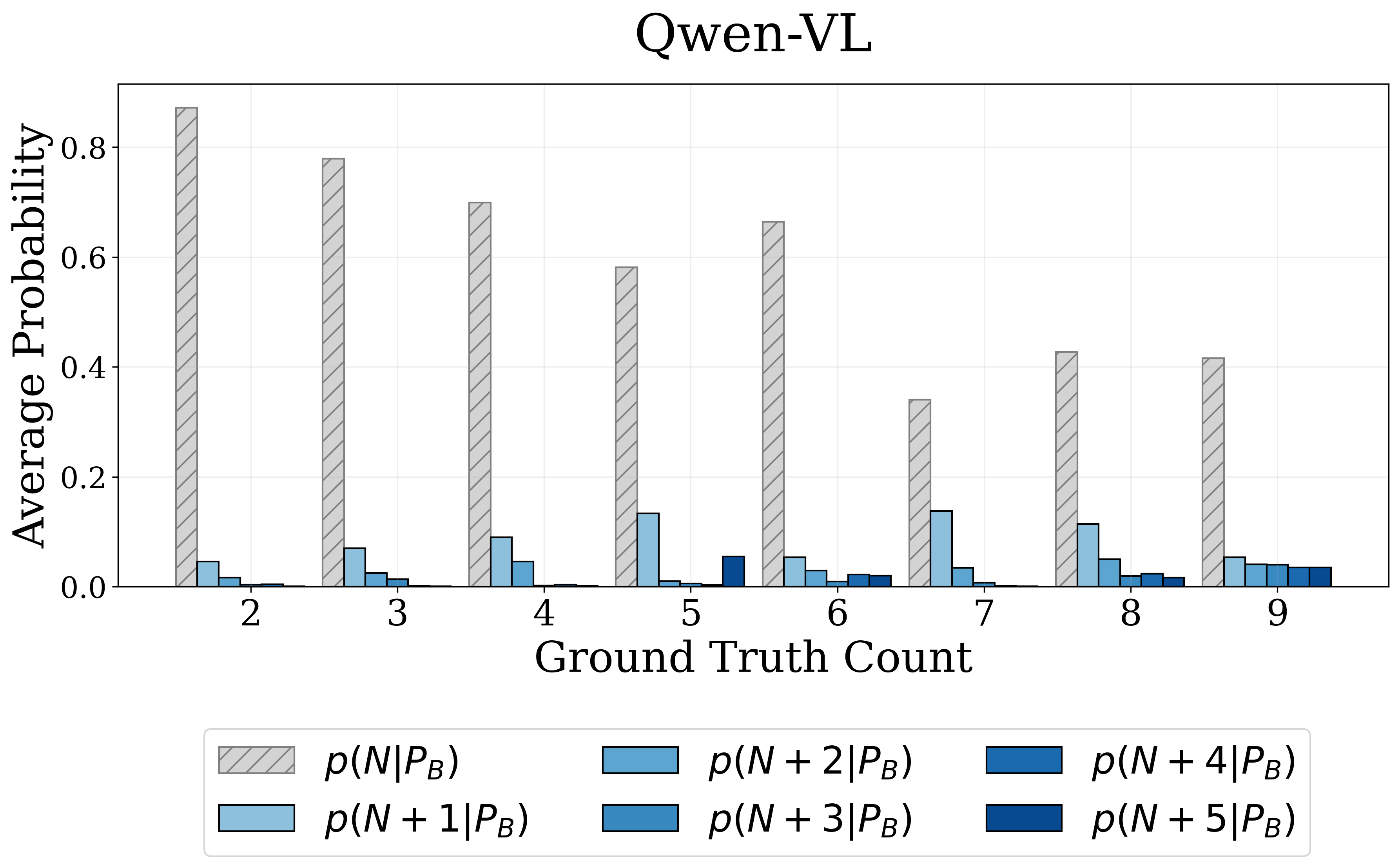}
    \caption{\textbf{Base probabilities on the object counting task.} We plot the probability assigned to the true count $N$ and neighboring counts $N+k$ ($k \in \{1,\ldots,5\}$) given the prompt ``How many [objects] are in the image?''. As object counts increase, confidence in $N$ decreases and probability mass spreads to neighboring counts.}
    \label{fig:base_probs}
\end{figure}

\subsection{PIH Increase as Confidence Decreases}
We hypothesize that models are more confident when counting fewer objects and therefore more resistant to misaligned prompts, with failures concentrated at minimal discrepancies ($k=1$).
We test this by collecting the probabilities for the true count $N$ and neighboring counts $N+k$ under the baseline prompt ($P_{B}$).
Figure~\ref{fig:base_probs} shows results for Qwen-VL, with similar trends for Janus-Pro and LlaVA-OneVision shown in \cref{app:base_probs_all}.
We filter out all samples for which $P_{B}$ is answered incorrectly, ensuring that the reported values reflect base confidence rather than raw accuracy.
Indeed, for $N \leq 4$, the probability $p(N \mid P_{B})$ is high while $p(N+k \mid P_{B})$ is low, indicating strong confidence in the predicted count.
At $N \geq 5$, base confidence drops and $p(N+k \mid P_{B})$ increases, matching our PIH findings.
Computing the Pearson correlation, we observe a moderately high $\rho$ between the base probability and confirmation to the prompt (computed per-sample), with $\rho = 0.37, 0.46$ for Qwen-VL, and Janus-Pro, respectively, as well as a small positive correlation of $0.1$ for LLaVA-OneVision.
Overall, these results demonstrate that base probabilities are correlated with model behavior under prompt-image mismatches, suggesting that PIH is more prevalent when visual confidence is low.

\section{Locating PIH Mechanisms}
\label{sec:mechanisms}
To better understand and reduce PIH, we aim to locate model components responsible for PIH.
Prior work has shown that attention heads often serve specific functional roles, such as information copying \cite{olsson2022context, feucht2025dual} and context tracking \cite{prakash2025language}, whereas MLP layers are more closely associated with storing parametric knowledge \cite{meng2023locating, arad2024refact}.
Therefore, we focus our analysis on identifying attention heads associated with PIH.

\subsection{Method: Attention Head Knockouts}
To analyze the causal role of individual attention heads in PIH, we perform attention head knockouts using mean ablation (\citet{wang2022interpretability, mueller2025mib}, see \cref{app:attention_knockout} for technical details).

Intuitively, this intervention removes token-specific information carried by the head while preserving its overall activation magnitude.
We apply this procedure head-by-head and quantify knockout success as the fraction of samples that switch from the prompted count to the true count under the misaligned prompt.
We rank heads by success and select the top-$m$ heads per model.
We then perform a second-stage grouped knockout in which all top-$m$ heads are ablated simultaneously for $m \in \{1,3,5,10\}$, selecting $m=3$ for Qwen-VL and $m=10$ for LLaVA-OneVision and Janus-Pro as a per-model hyperparameter. 
We refer to the final set of per-model attention heads as \textbf{PIH attention heads}.
See \cref{app:pih_heads} for a full list of top-$10$ heads per model and their success scores.

%Taking out for now...
% \begin{table}[t]
% \centering
% \small
% \begin{tabular}{l}
% \toprule
% \addlinespace[0.5em]
% \textbf{Qwen-VL} \\
% \addlinespace[0.5em]
% L0H3, L0H6, L11H15 \\
% \midrule
% \addlinespace[0.5em]
% \textbf{Llava-OneVision} \\
% \addlinespace[0.5em]
% L0H3, L0H6, L0H26, L11H15, L0H23, \\
% L0H24, L0H15, L14H9, L0H11, L17H22 \\
% \midrule
% \addlinespace[0.5em]
% \textbf{Jansu-Pro} \\
% \addlinespace[0.5em]
% L0H20, L1H7, L14H27, L12H9, L11H18, \\
% L13H2, L0H24, L14H28, L8H11, L8H3 \\
% \bottomrule
% \end{tabular}
% \caption{\textbf{Per-model PIH-heads.} Top attention heads per model, ranked by individual knockout success.
% We consider the number of top heads as a hyper-parameter, chosen per model.
% }
% \label{tab:pih_heads}
% \end{table}

\begin{table}[t]
\centering
\small
\begin{tabular}{lccc}
\toprule
 &
\textbf{LLaVA-OV} &
\textbf{Qwen-VL} &
\textbf{Janus-Pro} \\
\midrule

\addlinespace[0.5em]
\multicolumn{4}{l}{\textbf{Baseline Prompt: Exact Match} ($\uparrow$)} \\
Before & 76.89 & 78.49 & 80.32 \\
After (Random) & 77.30 & 78.70 & 80.00 \\
After (PIH) & 81.24 & 79.29 & 79.41 \\

\midrule
\addlinespace[0.5em]
\multicolumn{4}{l}{\textbf{Misaligned Prompt: Prompt Match ($\downarrow$)}} \\
Before & 42.58 & 56.51 & 64.10 \\
After (Random) & 37.80 & 54.60 & 58.30 \\
After (PIH) & 1.42  & 3.22 & 10.19 \\

\midrule
\addlinespace[0.5em]
\multicolumn{4}{l}{\textbf{Misaligned Prompt: True Count Match ($\uparrow$)}} \\
Before & 45.68 & 37.70 & 30.54 \\
After (Random) & 48.50 & 39.00 & 34.00 \\
After (PIH) & 77.80 & 70.66 & 70.90 \\

\bottomrule
\end{tabular}
\caption{\textbf{Response accuracies on baseline and misaligned prompts before and after ablations.}
The baseline prompt is ``How many [objects] are there in the image?''. Misaligned prompts take the form ``Describe the $N+k$ [objects] in the image'' ($k \in \{1,2,\ldots,5\}$) with results averaged across all $k$. 
As a random baseline, we sample an equal number of heads from the same layers as the PIH heads.
Lower prompt match indicates reduced PIH, while a higher true-count match reflects stronger correction toward visual evidence. Prompt and true-count matches do not sum to $100$\%, as some responses contain no numerical value.}
\vspace{-1em}
\label{tab:counting_knockouts_models}
\end{table}

\subsection{Attention Knockouts for PIH Mitigation}

\cref{tab:counting_knockouts_models} shows models' outputs given the baseline and misaligned prompts, before and after knockouts on the model-specific sets of PIH-heads. To verify that our results are specific to PIH heads rather than a general effect of ablation, we compare PIH heads to an equal number of randomly selected heads from the same layers in which the PIH heads reside.
First, our results show that ablating PIH-heads does not disrupt standard counting behavior, as demonstrated by the exact-match accuracies on the baseline prompt.
Notably, LLaVA-OneVision exhibits an improvement after ablation ($+4.35$ points), whereas Qwen-VL and Janus-Pro show minor changes of approximately $1$ point.
% Random head ablation is consistent with the $1$ point change in performance, and does not yield as big of a performance gain for LLaVA-OneVision as PIH head ablation.
Additionally, we verify that the object mentioned in the response remains unchanged and correct.
This confirms that ablating PIH-heads does not impact general counting capabilities.

Under the misaligned prompts, ablating PIH-heads substantially reduces prompt-following behavior, while random ablations produce only marginal reductions. 
Prompt match rates (previously corresponding to the blue curves in \cref{fig:confirmation}) drop from $42$-$64$\% before intervention to below $11$\% for all models after ablation.
Models also show a complementary increase in true count match rates, reaching $70$-$78$\% after intervention.
This shift indicates that the ablated heads are responsible for propagating the incorrect numerical information specified by the prompt.
Qualitative analysis of model generations before and after PIH-head ablation shows not only a reduction in incorrect numerical repetition but also the disappearance of spurious objects that were previously hallucinated and described in detail (examples are shown in \cref{app:generation_examples}).

Our findings show that PIH is causally mediated by a small set of attention heads that propagate tokens from the prompt.
Removing these heads shifts behavior from prompt reliance toward image reliance without harming general counting abilities, indicating that observed hallucinations are prompt-induced rather than due to counting failures. 

\begin{table}[t]
\centering
% \scriptsize
\small
\setlength{\tabcolsep}{3pt}
\begin{tabular}{lcccccc}
\toprule
 & \multicolumn{2}{c}{\textbf{CalTech101}} & \multicolumn{2}{c}{\textbf{MM-Vet}} & \multicolumn{2}{c}{\textbf{POPE}} \\
\cmidrule(lr){2-3} \cmidrule(lr){4-5} \cmidrule(lr){6-7}
\textbf{Model} & \textbf{Before} & \textbf{After} & \textbf{Before} & \textbf{After} & \textbf{Before} & \textbf{After} \\
\midrule
LLaVA-OV  & 100.0 & 100.0 & 50.5 & 50.7 & 89.3 & 89.3 \\
Qwen-VL   & 100.0 & 96.0  & 43.9 & 42.8 & 85.0 & 86.0 \\
Janus-Pro & 99.0  & 99.0  & 46.5 & 44.8 & 86.3 & 87.3 \\
\bottomrule
\end{tabular}
\caption{Accuracy on CalTech101, MM-Vet, and POPE before and after PIH-head ablation.}
\label{tab:mmvet_pope}
\end{table}

\subsection{Do PIH-Head Ablations Affect General Capabilities?}

While \cref{tab:counting_knockouts_models} provides initial evidence that PIH-head ablation does not degrade performance on CountBench, we further evaluate whether our intervention impacts broader capabilities. 

% \paragraph{Instruction Following and Copying.}
First, we test whether ablating PIH heads interferes with instruction following or general copying behavior. 
We construct a controlled task that does not introduce mismatches between the image and the prompt, in which copying is desired. Specifically, we adapt CalTech101 \cite{li2022caltech} and use the prompt:
\textit{``Repeat the following sentence if the image contains a(n) [OBJECT]. There is a(n) [OBJECT] in the image.''}. 
We measure whether the model correctly copies the second sentence conditioned on the image label.

% All models maintain near-perfect copying performance after ablation, with zero refusals observed. Minor deviations are due to benign corrections (e.g., ``beaver'' $\rightarrow$ ``otter'') or spelling differences, rather than failures of instruction following.

% \paragraph{Generalization Beyond Counting.}
We further evaluate models on MM-Vet \cite{yu2023mm} and POPE \cite{li2023evaluating} to assess broader multimodal capabilities. MM-Vet covers diverse skills such as recognition, OCR, spatial reasoning, and knowledge-based reasoning, while POPE primarily consists of existence-based questions (e.g., ``Is there a bottle in the image?''). Since these prompts are image-grounded and do not introduce explicit conflicts between the prompt and visual evidence, we do not expect PIH-head ablation to significantly affect performance.

% \begin{table}[h]
% \setlength{\tabcolsep}{8pt}
% \centering
% \small
% \begin{tabular}{lcc}
% \toprule
% \textbf{Model} & \textbf{Before} & \textbf{After (PIH)} \\
% \midrule
% LLaVA-OV  & 100.0 & 100.0 \\
% Qwen-VL   & 100.0 & 96.0  \\
% Janus-Pro & 99.0  & 99.0  \\
% \bottomrule
% \end{tabular}
% \caption{Copying accuracy (\%) on the CalTech101 conditional copying task.}
% \label{tab:copying}
% \end{table}

Performance remains broadly stable across all three benchmarks, with only minor fluctuations. This indicates that PIH-head ablation does not introduce negative side effects on general multimodal reasoning or standard image-grounded question answering.

\begin{table}[t]
\centering
\small
\begin{tabular}{ll}
\toprule
\textbf{Model} & \textbf{PIH Heads} \\
\midrule
LLaVA-OV &
\colorbox{blue!10}{\underline{L0H3}}, \colorbox{blue!10}{\underline{L0H6}}, \colorbox{blue!10}{L0H26}, \underline{L11H15}, \\
& \colorbox{blue!10}{L0H23}, \colorbox{blue!10}{\underline{L0H24}}, \colorbox{blue!10}{L0H15}, \colorbox{blue!10}{L14H9}, \\
& \colorbox{blue!10}{\underline{L0H11}}, \colorbox{blue!10}{L17H22} \\
\midrule
Qwen-VL &
\colorbox{blue!10}{\underline{L0H3}}, \colorbox{blue!10}{\underline{L0H6}}, \underline{L11H15}, \colorbox{blue!10}{\underline{L0H11}}, \\
& L15H8, L14H13, \colorbox{blue!10}{L0H10}, L7H3, \\
& \colorbox{blue!10}{\underline{L0H24}}, L8H13 \\
\midrule
Janus-Pro &
\colorbox{blue!10}{L0H20}, \colorbox{blue!10}{L1H7}, L14H27, L12H9, \\
& L11H18, L13H2, \colorbox{blue!10}{L0H24}, L14H28, \\
& L8H11, L8H3 \\
\bottomrule
\end{tabular}
\caption{Top-10 attention heads per model. Heads shared by Qwen-VL and LLaVA-OneVision are in \underline{underline}. Early-layer heads (L0-1) are \colorbox{blue!10}{highlighted}.}
\label{tab:pih_heads_all_models}
\end{table}

\subsection{Early-Layer LM Heads Mediate PIH}
% \cref{tab:pih_heads} shows the set of PIH-heads identified for each model.
We observe that PIH-heads exhibit a consistent pattern in their layer-wise localization.
Across all three models, top-$10$ PIH-heads (detailed in \cref{tab:pih_heads_all_models}) emerge in early and middle layers, suggesting that prompt-induced numerical information is introduced early and subsequently propagated forward.
In both Qwen-VL and LLaVA-OneVision, the top-ranked PIH-heads are L0H3 and L0H6, while in Janus-Pro the most influential head is likewise located in the first layer (L0H20).
This pattern extends beyond the top ranks: among the top-$10$ PIH-heads, $5/10$ occur in the $0$th layer for Qwen-VL, $7/10$ for LLaVA-OneVision, and $3/10$ fall within the first two layers (layers $0$–$1$) for Janus-Pro. 
The early-layer concentration of PIH-heads, highlighted in \cref{tab:pih_heads_all_models}, suggests that numerical information from the prompt is injected before extensive multimodal integration or higher-level reasoning takes place, enabling it to shape downstream computation even when it conflicts with visual evidence.

This early-layer localization raises a related question about the source of these mechanisms: namely, whether PIH arises from vision–language interaction or from the language model itself.
One way to interpret this question is through the overlap of PIH-heads between Qwen-VL and LLaVA-OneVision, which share the same underlying Qwen2 language model but use different vision backbones and image–language fusion architectures. % (Qwen2-VL uses a dynamically tokenized ViT with single-stream fusion, while LLaVA-OneVision uses a SigLIP-style ViT with projector-based fusion

The top-ranked PIH-heads exhibit substantial overlap across the two models, underlined in \cref{tab:pih_heads_all_models}: the top-1 and top-2 heads overlap perfectly, and at $m=10$, half of the identified heads are shared.
This level of overlap strongly suggests that PIH is primarily mediated by language-model attention heads, rather than arising from vision-specific components or cross-modal alignment layers. 
This is consistent with prior work showing that fine-tuning tends to amplify or repurpose existing mechanisms rather than introduce entirely new mechanisms \citep[e.g.,][]{prakash2024fine}. Further, this supports the view that PIH reflects language-internal information routing that persists across multimodal models.

% \begin{table}[t]
% \centering
% \small
% \begin{tabular}{lccc}
% \toprule
% \textbf{Condition} &
% \textbf{LLaVA-OV} &
% \textbf{Qwen-VL} &
% \textbf{Janus-Pro} \\
% \midrule

% \addlinespace[0.5em]
% \multicolumn{4}{l}{\textbf{Base Prompt: Exact Match} ($\uparrow$)} \\
% Before & 96.25 & 96.25 & 96.88 \\
% After  & 95.97 & 95.60 & 96.23 \\

% \midrule
% \addlinespace[0.5em]
% \multicolumn{4}{l}{\textbf{Misaligned Prompt: Prompt Match ($\downarrow$)}} \\
% Before & 62.09 & 20.04 & 74.95 \\
% After  & 3.07  & 12.72 & 41.01 \\

% \midrule
% \addlinespace[0.5em]
% \multicolumn{4}{l}{\textbf{Misaligned Prompt: True Color Match ($\uparrow$)}} \\
% Before & 37.91 & 79.96 & 25.05 \\
% After  & 96.93 & 87.28 & 58.99 \\

% \bottomrule
% \end{tabular}
% \caption{Color classification accuracy under baseline and misaligned prompts.
% The \emph{baseline prompt} asks, ``What color is the [object]?''.
% The \emph{misaligned prompt} asks the model to ``Describe the $C+k$ [object],''
% where $C$ denotes the ground-truth color of the object in the image and
% $k \in \{1,2,3\}$ denotes the perceptual difference from $C$ on the color wheel.
% Lower color-bias match ($\downarrow$) indicates fewer prompt-induced color hallucinations,
% while higher ground-truth color match ($\uparrow$) reflects stronger correction toward the visual evidence.}
% \label{tab:color_knockouts_models}
% \end{table}

% \begin{enumerate}
%     \item Add any additional tasks. 
% \end{enumerate}

\begin{table*}[t]
\setlength{\tabcolsep}{12pt}
\centering
\small
\begin{tabular}{lcccccc}
\toprule
& \multicolumn{2}{c}{\textbf{LLaVA-OV}} 
& \multicolumn{2}{c}{\textbf{Janus-Pro}} 
& \multicolumn{2}{c}{\textbf{Qwen-VL}} \\
\cmidrule(lr){2-3} \cmidrule(lr){4-5} \cmidrule(lr){6-7}
\textbf{Response}
& \textbf{Before} & \textbf{After}
& \textbf{Before} & \textbf{After}
& \textbf{Before} & \textbf{After} \\
\midrule
NC (correct)        & 0.96  & 46.01 & 5.17  & 20.94 & 8.93  & 30.24 \\
FC (correct)        & 0.00  & 33.23 & 9.61  & 24.14 & 11.34 & 39.86 \\
NC (no color)       & 0.00  & 15.97 & 0.00  & 10.34 & 0.00  & 9.62  \\
\midrule
\textbf{No PIH}
& 0.96  & \textbf{95.21} & 14.78 & \textbf{55.42} & 20.27 & \textbf{79.72} \\
\midrule
EC (incorrect)      & 93.61 & 3.51  & 82.76 & 41.13 & 28.87 & 14.09 \\
SC (incorrect)      & 5.43  & 1.28  & 2.46  & 3.45  & 50.86 & 6.19  \\
\midrule
\textbf{PIH}
& \textbf{99.04} & 4.79  & \textbf{85.22} & 44.58 & \textbf{79.73} & 20.28 \\
\bottomrule
\end{tabular}
\caption{\textbf{Distribution of PIH and no PIH cases before and after PIH attention head ablation for the color task and semantic response types (\%).}
NC, FC, SC, and EC denote no-, format-, soft-, and exact-copying, respectively.
Correct denotes the ground-truth image color, while incorrect denotes the color in the misaligned prompt.
Summed rows highlight the shift from PIH to image-grounded behavior after ablation.}
\label{tab:semantic_response_comparison}
\end{table*}

\section{Generalization Beyond Counting}
\label{sec:pih_generalization}

Given that PIH appears to be mediated by shared language-model attention heads that persist across multimodal architectures, we next ask whether these same heads generalize beyond counting to mitigate prompt-induced hallucinations in other tasks.
Specifically, we use a color prediction task based on the Visual CounterFact dataset \cite{pvp}, as color naturally allows gradual discrepancies.
We use ''What color is the [object] in the image?'' as the baseline prompt, and ``Describe the $C+k$ [object]'' as the misaligned prompt, where $C$ denotes the ground-truth color of the object and $k \in \{1,2,3\}$ denotes the perceptual ``difference'' from the $C$ on the color wheel.
Additional details on the color task are available in \cref{app:colors}. 
To test generalization, we use the same set of PIH-heads per model (\cref{tab:counting_knockouts_models}) on the color task. 
Table~\ref{tab:semantic_response_comparison} show that ablating the same PIH heads yields large improvements under misaligned color prompts across all models, reducing prompt-induced hallucinations by between \textit{40\% and 95\%}. This demonstrates that PIH heads identified through counting generalize beyond numerical reasoning and are not task-specific.
% For LLaVA-OV, accuracy increases from $0.96\%$ to $95.21\%$, while Janus-Pro and Qwen-VL improve from $14.78\%\!\rightarrow\!55.42\%$ and $20.27\%\!\rightarrow\!79.72\%$, respectively.
% These gains demonstrate that PIH head ablation effectively reduces prompt-induced hallucinations in a different domain despite the heads being identified using numerical supervision alone.}

\section{Analyzing the Function of PIH-heads}
% Currently, we only include the plot with P(N_digit| N_digit). I can quickly run P(N_word | N_word) -- so instead of "describe the 2 cats' run 'describe the two cats'. This would run fast since I'd only test exactly  P(N_wor--- At this point, we run for rebuttals only... 

\label{sec:pih_functions}
In this section, we analyze PIH-heads to better understand their functionality.
We hypothesize that ablating PIH-heads induces one of two mechanisms: PIH-heads could reduce hallucinations by inhibiting copying behavior or by increasing attention flow to the image. Namely, PIH-heads could enable the model to propagate incorrect information from the prompt without referencing the image. Alternatively, PIH-heads may influence hallucinations by modulating the model's reliance on the image, so that removing these heads increases reliance on visual evidence.

To determine the functionality of PIH heads, we first investigate their attention patterns over the input prompts and find that they do not reveal a consistent pattern connecting the generated answer to either copying the prompt or relying on the image (Examples shown in \cref{app:attention_patterns}).
This is in line with several studies showing that influential attention heads may not exhibit human-interpretable patterns \cite{jain2019attention, serrano2019attention, brunner2019identifiability, grimsley2020attention}. 
Therefore, we study the function of PIH-heads by examining the generated texts before and after knockouts, shifts in attention mass after PIH-ablation, and the impact of PIH-ablation on model confidence.
Through these analyses, we show that PIH ablation reduces hallucinations by both \textbf{inhibiting copying} of information from the prompt and \textbf{increasing attention} to the image. 

\begin{table}[t]
\centering
\resizebox{\columnwidth}{!}{%
\small
\setlength{\tabcolsep}{4pt}
\renewcommand{\arraystretch}{1.15}
\begin{tabular}{ll}
\toprule
\multicolumn{2}{l}{\textbf{Prompt: ``Describe the 3 cats.'' (N=2)}} \\
\midrule
\hspace{1em} \textbf{Response} & \hspace{3em} \textbf{Copying Form} \\
\midrule
\hspace{1em} ``There are 3 cats.'' & \hspace{3em} Exact copying \\
\hspace{1em} ``There are three cats.'' & \hspace{3em} Soft copying \\
\hspace{1em} ``There are 2 cats.'' & \hspace{3em} Format copying \\
\hspace{1em} ``There are two cats.'' & \hspace{3em} No copying \\
\midrule
\multicolumn{2}{l}{\textbf{Prompt: ``Describe the purple banana.'' (N=yellow)}} \\
\midrule
\hspace{1em} \textbf{Response} & \hspace{3em} \textbf{Copying Form} \\
\midrule
\hspace{1em} ``The purple banana \ldots'' & \hspace{3em} Exact copying \\
\hspace{1em} ``The banana is purple.'' & \hspace{3em} Soft copying \\
\hspace{1em} ``The yellow banana \ldots'' & \hspace{3em} Format copying \\
\hspace{1em} ``The banana is yellow.'' & \hspace{3em} No copying \\
\bottomrule
\end{tabular}}
\caption{\textbf{Definition of copying forms under misaligned prompts for count and color tasks.}}
\label{tab:copying_definitions}
\end{table}

\subsection{Ablation Impacts Copying Form}

\begin{table}[t]
\centering
\small
\setlength{\tabcolsep}{4pt}
\renewcommand{\arraystretch}{1.15}
\begin{tabular}{llcccc}
\toprule
\textbf{Model} & \textbf{Accuracy} &
\multicolumn{2}{c}{\textbf{Digit (\%)}} &
\multicolumn{2}{c}{\textbf{Word (\%)}} \\
\cmidrule(lr){3-4}\cmidrule(lr){5-6}
 &  & Before & After & Before & After \\

\midrule
\multirow{3}{*}{LLaVA-OV}
& Incorrect & 7.57  & 0.25 & 45.62 & 15.29 \\
& Correct   & 0.30  & 0.25 & 46.51 & 84.21 \\
& \textbf{Total} & \textbf{7.87} & \textbf{0.50} & \textbf{92.13} & \textbf{99.50} \\

\midrule
\multirow{3}{*}{Qwen-VL}
& Incorrect & 28.44 & 4.53 & 33.11 & 16.94 \\
& Correct   & 0.05  & 0.00 & 38.40 & 78.53 \\
& \textbf{Total} & \textbf{28.49} & \textbf{4.53} & \textbf{71.51} & \textbf{95.47} \\

\midrule

\multirow{3}{*}{Janus-Pro}
& Incorrect & 43.45 & 2.67 & 25.40 & 23.30 \\
& Correct   & 0.50  & 0.14 & 30.65 & 73.89 \\
& \textbf{Total} & \textbf{43.95} & \textbf{2.81} & \textbf{56.05} & \textbf{97.19} \\
\bottomrule
\end{tabular}
\caption{\textbf{Response format and correctness before and after PIH attention head ablation.}
Digit responses indicate copying format used in the prompt, while word responses correspond to spelled-out number expressions.}
\label{tab:number_format_hybrid}
\end{table}

To characterize how models incorporate misaligned prompt information, we categorize several forms of copying behavior, summarized in Table~\ref{tab:copying_definitions}. 
\textit{Soft-copying} occurs when the response contains the misaligned prompt information but deviates in structure from the prompt's structure.
\textit{Format copying} occurs when the response contains the correct (visually grounded) information, while maintaining the same prompt format.
\textit{Exact copying} occurs when both the prompt information and the format are present.
Using these definitions, we analyze copying behavior first in the counting task and then in the color recognition task.

Table~\ref{tab:number_format_hybrid} shows copying patterns in the counting task, where the input prompts always include the number as a digit (i.e., ``Describe the $N_{\text{digit}}$ [objects]''). 
Prior to ablation, responses that include the number as a digit are more likely to be incorrect. For example, Janus-Pro outputs a digit in $43.95$\% of the cases, consisting of $43.45$\% of the incorrect cases, and $0.05$\% correct cases. This indicates that incorrect responses stem from exact copying of misaligned prompt information. 
After ablating PIH-heads, digit responses are almost entirely eliminated across all models. Notably, the reduction in digit responses occurs for both incorrect and correct outputs, showing that the suppression of format copying is not merely a byproduct of improved numerical accuracy.

% additional nums
% For Janus-Pro, total digit usage drops from $43.45\%$ to $2.81\%$, while word-based responses increase correspondingly to $97.19\%$, with similar trends for LLaVA-OneVision and Qwen-VL.
% Similar shifts are observed for LLaVA-OneVision ($7.87\%\!\rightarrow\!0.50\%$ digit) and Qwen-VL ($28.49\%\!\rightarrow\!4.53\%$ digit). 

The different copying forms in Table~\ref{tab:semantic_response_comparison} show that PIH in the color task is likewise associated with copying the format of the prompt. 
Before ablation, models frequently reproduce the prompt’s structure through format copying responses, accounting for over $90\%$ of outputs for both LLaVA-OneVision and Janus-Pro.
After PIH head ablation, this behavior is substantially reduced and accompanied by a corresponding increase in free-form descriptions.
% additional num
% Namely, format copying responses drop from $93.61\%\!\rightarrow\!36.74\%$ for LLaVA-OneVision and from $92.37\%\!\rightarrow\!65.27\%$ for Janus-Pro, accompanied by a corresponding increase in free-form descriptions. 

These results suggest that when PIH-heads are active, models tend to reproduce both the content and the structure of misaligned prompts; when they are ablated, responses shift toward freer, less prompt-constrained descriptions that better reflect visual evidence. This parallel between counting and color tasks indicates that PIH is not limited to numerical reasoning, but reflects a broader tendency to copy information from the prompt, even when it is inconsistent with the image.

\subsection{PIH-Ablation Shifts Attention Mass to Image Tokens}
\begin{figure}[t]
    \centering
    \includegraphics[width=0.85\linewidth]{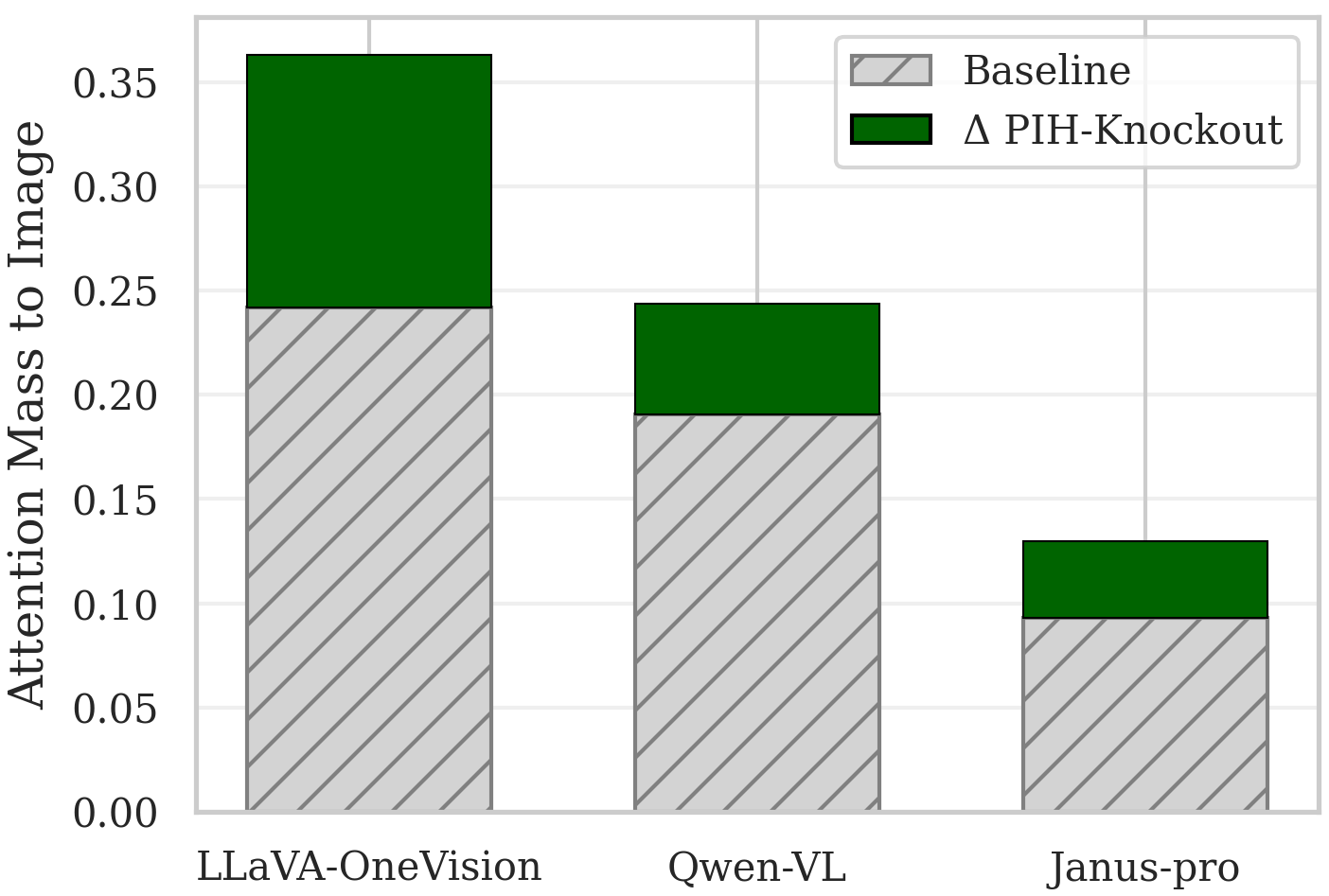}
    \caption{Image attention mass at the layer with the largest intervention-induced change for each model. Namely, LLaVA-OneVision (layer~2, $\Delta$=$0.121$), Qwen-VL (layer~27, $\Delta$=$0.053$),and Janus-pro (layer~22, $\Delta$=$0.037$).
        Gray bars indicate the baseline attention mass for image tokens, while green bars show an increase after PIH-knockout. 
        %LLaVA-OneVision shows both the highest baseline image attention and the strongest intervention effect, indicating a greater reliance on visual tokens.
    }
    \label{fig:max_delta_layer_image_attention}
\end{figure}
%For reference, the most sensitive layers and corresponding attention changes are:LLaVA-OneVision (layer~2, $\Delta=0.121$), Qwen-VL (layer~27, $\Delta=0.053$),and Janus-pro (layer~22, $\Delta=0.037$).

One notable outcome of PIH head ablation is observed in LLaVA-OneVision, where performance improves even for non-conflicting prompt–image cases. 
As shown in \cref{tab:counting_knockouts_models}, baseline counting accuracy increases by $4.35\%$, despite the absence of misleading prompt information. Similarly, in the color task, PIH ablation yields a strong reduction in prompt-copying, with PIH reduced by $94\%$. These gains suggest that PIH ablation affects more than prompt-following, potentially altering how models balance textual and visual inputs.

To investigate this effect, we analyze changes in \emph{attention mass} from assistant tokens to image tokens. For each layer, we compute the fraction of attention allocated to image tokens, with the remainder allocated to text tokens (summing to $1$). 
\cref{fig:max_delta_layer_image_attention} shows the layer with the largest shift in attention from text to image, after PIH ablation: layer $2$ in LLaVA-OneVision, $22$ in Qwen-VL, and $27$ in Janus-Pro; see \cref{app:attention_mass} for all layers.
PIH head ablation increases image attention across all models, most strongly in LLaVA-OneVision ($\Delta = 0.121$) which shows an increase in an early model layer. 
This is significant, as previous work has shown the importance of visual information processing in early layer for multimodal integration \cite{nikankin2025same}.
These indicate that, particularly in LLaVA-OneVision, PIH head ablation induces a reallocation of attention toward visual inputs, providing a plausible explanation for the observed improvements in both baseline counting accuracy and robustness to PIH.

% additional nums
% PIH head ablation significantly increases attention to image tokens, with the magnitude of this shift varying across models. 
% LLaVA-OneVision exhibits both the highest baseline image attention and the largest increase after ablation ($\Delta = 0.121$), compared to more modest changes in Qwen-VL ($\Delta = 0.053$) and Janus-Pro ($\Delta = 0.037$). 

\subsection{PIH Heads Mediate Model-Specific Copying Mechanisms}
\begin{figure}
    \centering
    \includegraphics[width=0.9\columnwidth]{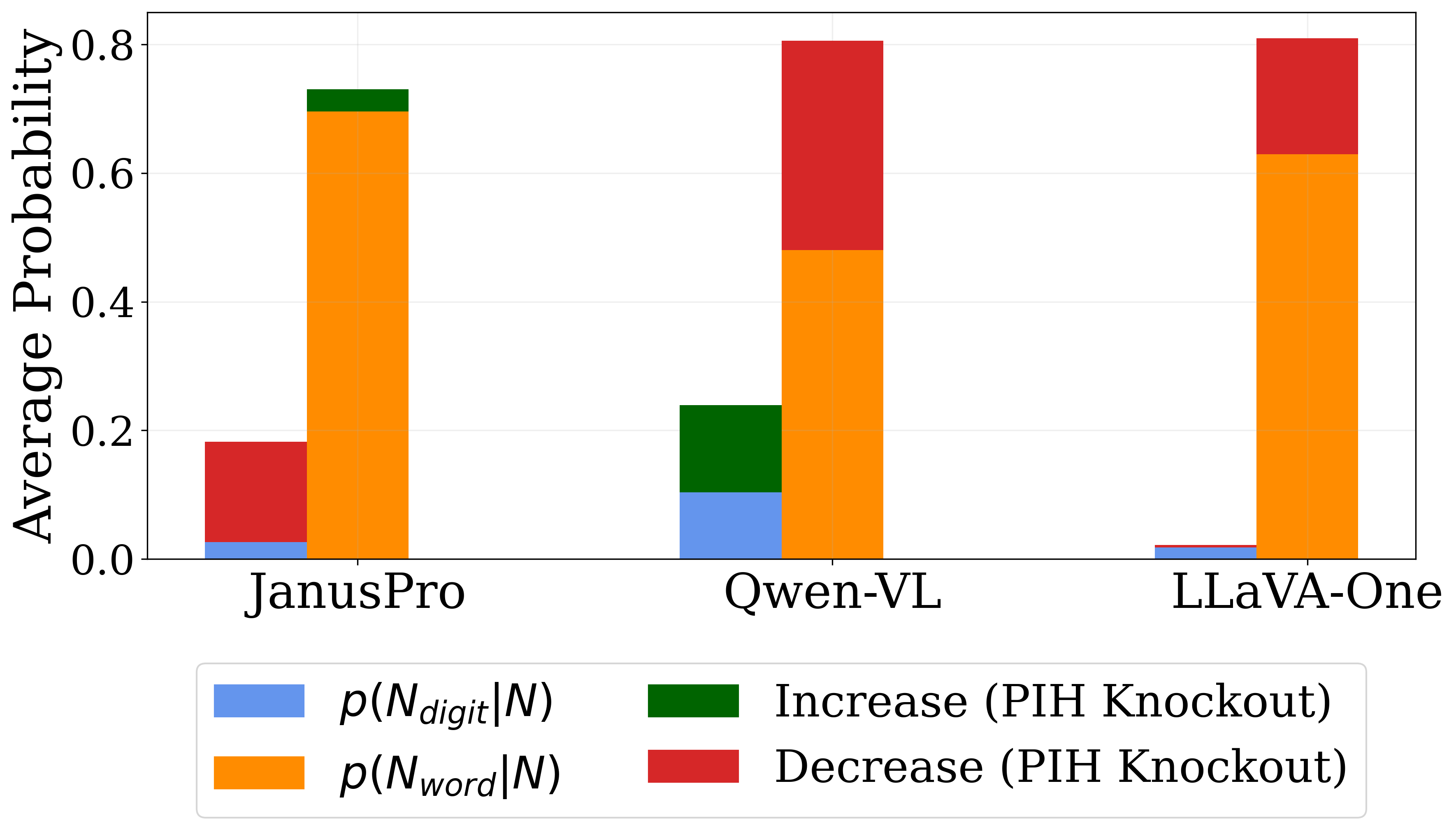}
    \caption{
    \textbf{Impact of knockout on the probability of the correct count in digit (blue) and word form (orange)}. We measure the probabilities given the prompt ``Describe the $N_{\text{digit}}$ [objects]''. The knockout effects are shown in green (increase) and red (decrease).
    }
    \label{fig:knockout_prob_impact}
\end{figure}

Beyond analyzing changes in model outputs, we seek to understand how PIH head ablation affects ground-truth probabilities when presented with prompts that do not conflict with the image.
We examine the conditional probability of the ground-truth count $N$ when expressed as either a digit ($N_{\text{digit}}$) or a word ($N_{\text{word}}$) under prompts of the form ``Describe the $N_{\text{digit}}$ [objects]'', shown in \cref{fig:knockout_prob_impact}. We use a digit-word analysis because it offers a controlled way to distinguish format-copying from soft-copying by analyzing a single token.
%In particular, if $P(N_{\text{digit}} \mid \text{Describe} \ N_{\text{digit}})$ decrease, then ablation suppresses format copying and if  $P(N_{\text{word}} \mid \text{Describe} \ N_{\text{digit}})$ decreases then ablation inhibits soft-copying. 

\paragraph{Janus-Pro inhibits format copying when correcting misaligned prompts.} \cref{fig:knockout_prob_impact} demonstrates that $P(N_{\text{digit}} \mid N_{\text{digit}})$ significantly decreases when ablating PIH-Heads, which indicates a reduction in format-copying in the counting task. This is supported by the finding that PIH ablation in Janus-Pro reduces format copying by nearly $30\%$, a shift that is consistently accompanied by an increase in correct, free-form responses.
While $P(N_{\text{digit}} \mid N_{\text{digit}})$ decreases, PIH ablation increases $P(N_{\text{word}} \mid N_{\text{digit}})$, indicating that when correcting misaligned prompts, Janus-Pro more frequently produces the ground-truth response in a word-based, free-form format rather than copying the digit form used in the prompt. 
Additionally, our findings show that PIH ablation produces only a modest shift in attention mass toward the image, consistent with almost no change in baseline counting performance.
Together, these results suggest that PIH ablation in Janus-Pro does not increase visual reliance but primarily inhibits prompt-based format copying when correcting misaligned prompts.

% Additional Stats
 %In particular, PIH-ablation increases the $P(N_{\text{word}} \mid N_{\text{digit}})$ for count and free-form responses containing the ground truth-color increase by $~27\%$. 
%Figure~\ref{fig:max_delta_layer_image_attention} shows that PIH-ablation provides a slight increase in attention mass; however, this increase in reliance on the image is not enough to improve performance on the base-counting prompt. The decrease in $P(N_{\text{digit}} \mid N_{\text{digit}})$ and the reduced rate of format copying in the color task provide complementary evidence that PIH ablation suppresses format copying while encouraging alignment with the ground-truth image content.

\paragraph{Qwen-VL encourages format copying when correcting misaligned prompts.} In contrast to Janus-Pro, we find that PIH-ablation causes Qwen-VL to more heavily rely on the prompt \textit{format}. Figure~\ref{fig:knockout_prob_impact} shows a $32.6\%$ decrease in probability of $P(N_{\text{word}} \mid N_{\text{digit}})$ with a corresponding \textit{increase} in corrected responses that follow the digit format of the prompt. Namely, $P(N_{\text{digit}} \mid N_{\text{digit}})$ increases by $13.6\%$. This is consistent with Table~\ref{tab:semantic_response_comparison}, which shows that Qwen-VL is the only model for which PIH head ablation increases format copying, with copying rates rising from $40.21\%$ to $53.95\%$. While PIH ablation increases image attention and improves correction performance in Qwen-VL,
Figure~\ref{fig:knockout_prob_impact} reveals that these gains arise from a shift away from soft copying toward increased format copying, rather than a strict reduction in copying.

% additional evidence
%Figure~\ref{fig:max_delta_layer_image_attention} shows that PIH ablation increases reliance on the image, and Tables~\ref{tab:acc_by_count} and ~\ref{tab:color_accuracy_only} confirm that knockout improves correction performance in Qwen-VL. However, Figure~\ref{fig:knockout_prob_impact} reveals that these gains arise not from reduced copying overall, but from a shift away from soft-copying toward increased format copying, indicating that PIH ablation in Qwen-VL amplifies prompt-format dependence rather than suppressing it.

\paragraph{LLaVA-OneVision inhibits all forms of copying and shows a greater reliance on the image when correcting misaligned prompts.} Figure~\ref{fig:knockout_prob_impact} and Table~\ref{tab:number_format_hybrid} show that LLaVA-OneVision rarely produces digit-form responses even prior to PIH ablation, indicating format copying suppression.
Next, Figure~\ref{fig:knockout_prob_impact} demonstrates that$P(N_{\text{word}} \mid N_{\text{digit}})$ heavily decreases ($~18\%$), indicating LLaVA-OneVision also suppresses soft copying.
In terms of image reliance, LLaVA-OneVision is the only model that, after PIH ablations, showed a substantial increase in performance on the baseline counting prompt, indicating an improvement in general counting abilities.
In parallel, we find that PIH-ablation leads to the largest shift in attention mass toward the image ($+12\%$), indicating a greater reliance on the image.

Although implemented differently across models, PIH head ablation consistently reduces prompt copying and increases visual reliance, highlighting the promise of targeted interventions.

% additional evidence
% \cref{fig:knockout_prob_impact} shows that across all models, knockout successfully increases the probability of the ground truth count when provided with a modality conflict in the prompt, indicated by the blue bars with an increase shown in green.
% However, the probability of $N$ conditioned on ''Describe the $N$ [objects]'' decreases after knockout, indicating a reduction in copying behavior. This trend of decreasing $P(N \mid N)$ is observed across all models and tasks, except for LLaVA-OV on color. In this case, we observe a slight increase in the probability of predicting the correct color when PIH ablation is performed. Recall that for LLaVA-OV, Table~\ref{tab:color_accuracy_plus_format} shows we see the largest increase of correcting to the correct color \textit{and} removing format copying. 

\section{Conclusion}
In this work, we study prompt-induced hallucinations in VLMs through a controlled object-counting task, isolating text-image conflicts. Through mechanistic analysis of three models, we show that such hallucinations stem from systematic prompt-based copying rather than failures of visual perception, and can be substantially reduced by ablating a small set of attention heads.
We find that PIH is localized primarily to early layers of the language model, where prompt-based numerical information is likely processed and propagated before multimodal integration. 
Additionally, we find that Qwen-VL and LLaVA-OneVision share many of these influential heads, indicating they largely originate from the shared language-model backbone rather than vision-specific components.

Ablating PIH heads consistently shifts behavior toward visually grounded corrections, generalizes beyond counting to color prediction, and does not degrade general performance.
Finally, we find that PIH heads mediate different forms of prompt copying with model-specific implementations.
LLaVA-OneVision exhibits the strongest effects, including broad suppression of all forms of prompt-copying and substantially increased reliance on visual information after PIH head ablations. 
Our results highlight PIH as a mechanistically localized and model-dependent failure mode in VLMs, suggesting targeted interventions can improve visual grounding under prompt–image mismatches.

\section*{Limitations}
Our study focuses on medium-scale VLMs ($\sim$7B parameters). While these models are representative of widely used multimodal architectures, the specific PIH heads identified here and the effects of their ablation may not generalize directly to substantially larger models.

Furthermore, our analysis relies on attention-based methods, which offer only a partial view of internal computation. 
As noted in prior work, attention patterns are often not directly interpretable \cite{jain2019attention, serrano2019attention, brunner2019identifiability, grimsley2020attention}, and heads with strong behavioral influence may not exhibit clear or human-aligned attention behaviors.
Consistent with this, we do not observe simple or consistent attention patterns within PIH heads themselves.

Further, ablating PIH heads may induce secondary changes in the behavior of other attention heads or layers. These secondary effects on other attention heads may be more human-interpretable than the original PIH heads. We do not explicitly trace how these secondary changes emerge in this work, and leave their detailed analysis to future studies.

While we demonstrate all models follow the same general trend of a reduction in exact copying and an increase in attention mass to the image, we find that the manner in which copying is reduced is distinct for all three models. Determining the exact mechanism as to why PIH ablations causes a reduction in soft copying, while encouraging format copying in Qwen-VL and results in a strict decrease in format copying with an increase of soft copying Janus-Pro, is left for future works. Developing a better understanding the exact reason, whether its due to differences in architecture, patterns in selected heads or training data would help to give a more complete picture of prompt-induced hallucinations and the model components that mediate them.

\section*{Acknowledgments}
This research was partly funded by the European Union (ERC, Control-LM, 101165402). Views and opinions expressed are however those of the author(s) only and do not necessarily reflect those of the European Union or the European Research Council Executive Agency. Neither the European Union nor the granting authority can be held responsible for them.
YB and DA were partially supported by the Israel Science Foundation (grant no.\ 2942/25) and by Coefficient Giving.
DA is supported by the Ariane de Rothschild Women Doctoral Program. K.M. and W.R. were supported by NSF grant 2313027. This work was partly supported by the National Science Foundation under Cooperative
Agreement 2421782 and the Simons Foundation award MPS-AI-00010515
(NSF-Simons AI Institute for Cosmic Origins - CosmicAI, https://www.cosmicai.org/

\bibliography{bib}
\appendix

\section{Experimental Details}
\label{app:exp_setup}

\begin{table*}[h]
    \centering
    \small
    \resizebox{\textwidth}{!}{
    \begin{tabular}{lcccc}
        \toprule
        \textbf{Model Name} & \textbf{Underlying LM} & \textbf{Vision Encoder} & \textbf{Size} & \textbf{HuggingFace Path} \\
        \midrule
        LLaVA-OneVision & Qwen2 & SigLIP & 7B & \texttt{llava-hf/llava-onevision-qwen2-7b-ov-hf} \\
        Qwen-VL & Qwen2 & DFN-ViT w/ RoPE-2D & 7B & \texttt{Qwen/Qwen2-VL-7B-Instruct} \\
        Janus Pro & DeepSeek-LLM & SigLIP-Large-Patch16-384 & 7B & \texttt{deepseek-ai/Janus-Pro-7B} \\
        \bottomrule
    \end{tabular}
    }
    \caption{Details of evaluated models, including underlying language models, vision encoders, sizes, and Hugging Face model paths.}
    \label{tab:model_details}
\end{table*}

\subsection{Computational Requirements}

All experiments in this work were conducted using a single NVIDIA RTX~3090 GPU with 24~GB of memory.
Our analysis does not involve model training or fine-tuning, and instead relies on forward inference, attention-head ablation, and probability extraction.
The complete set of experiments, including exploratory analyses and ablations not included in the final paper, consumed approximately 200-300 GPU hours.
Peak memory usage occurred during attention output analysis, but all experiments fit within a single GPU.

% \section{Hardware Details}
% \label{app:hardware}
% ll experiments were conducted on a single workstation equipped with an NVIDIA RTX~3090 GPU with 24~GB of VRAM.
% The system was used for inference-only analysis, including attention-head ablation, probability extraction, and attention mass computation.
% No multi-GPU parallelism or distributed computation was required.

\subsection{Licenses and Third-Party Usage}
\label{app:licenses}
This work is implemented using PyTorch \citep{paszke2019pytorch}, an open-source deep learning framework licensed under the BSD license, and the Hugging Face Transformers library \citep{wolf2019huggingface}, licensed under Apache 2.0. All software usage complies with their respective license terms.
For evaluation, we use CountBench \cite{paiss2023teaching} and Visual CounterFact \cite{pvp} licensed under the Apache 2.0 license.

\subsection{Models}
% TODO: Did you report the number of parameters in the models used. try to have a nice tabel w/ all the diff vision encoders/sizes.
We evaluate a three diverse VLMs. Table~\ref{tab:model_details} provides details on the open-source models used in our experiments. These models are sourced from Hugging Face, with their specific repository paths listed for reproducibility.

\subsection{Datasets}
We use CountBench \cite{paiss2023teaching} for our counting task and Visual CounterFact \cite{pvp} for our color identification task. CountBench consists of 491 text-image pairs. We adapt CountBench using our novel misaligned prompts, yielding a total of 3,437 image-prompt pairs. We use the ``color'' split of the Visual CounterFact dataset, which comprises 493 base-image-label pairs. We adapt Visual CounterFact with our base and misaligned prompts, yielding 2,465 image-prompt pairs.

\subsection{Generation Details}
For each image with ground-truth object count $N$, we first present the model with a \emph{baseline prompt}, $P_{B} = $ ``How many [objects] are there in the image?'', and retain only samples for which the model correctly predicts $N$.
We then introduce \emph{misaligned prompts} asking the model to ``Describe the $N+k$ [objects] in the image,'', where $k \in \{1,2,\ldots,5\}$. 
Here, the offset $k$ serves as a \emph{discrepancy distance}, measuring the magnitude of misalignment between the prompt and the visual input.

We allow the model to generate long-form responses up to $75$ tokens and use a rule-based algorithm to determine accuracy by extracting the first non-negated numerical reference in the response (e.g., excluding constructions such as ``there are not eight waterlilies'').
A PIH occurs when the model proceeds to describe $N+k$ objects without correcting the count, despite contradictory visual evidence.

\subsection{Accuracy on CountBench}
We first evaluate the baseline accuracy of LlaVA-OneVision, Qwen-VL and Janus-Pro on CountBench \cite{paiss2023teaching}. Table~\ref{tab:acc_by_count} breaks performance down by the number of objects. All models perform reliably at very small counts (2–4 objects), but accuracy declines as counts increase. Notably, there are sharp dips at \textbf{5 objects} (e.g., LLaVA-1.6: 87.27\% to 65.91\%, QwenVL: 90.91\% to 75.00\%) and again at \textbf{7 objects} (Janus-Pro: 87.72\% to 55.56\%, LLaVA-OneVision: 89.47\% to 48.15\%). These discontinuities suggest the presence of training biases that systematically disadvantage certain numbers, a phenomenon supported by prior work linking VLM counting performance to biases in object frequency distributions seen during pretraining \cite{rudman2025forgotten, vo2025vision}. 

\begin{table}[h]
\centering
\begin{tabular}{lccc}
\hline
\textbf{Num} & \textbf{LLaVA-OV} & \textbf{Janus-Pro} & \textbf{QwenVL} \\
\hline
2  & 90.57 & 84.91 & 90.57 \\
3  & 91.23 & 87.72 & 85.96 \\
4  & 87.27 & 85.45 & 90.91 \\
5  & 65.91 & 70.45 & 75.00 \\
6  & 89.47 & 87.72 & 85.96 \\
7  & 48.15 & 55.56 & 51.85 \\
8  & 71.93 & 80.70 & 75.44 \\
9  & 68.33 & 86.67 & 71.67 \\
\hline
\end{tabular}
\caption{Performance across different numbers of objects on CountBench.}
\label{tab:acc_by_count}
\end{table}

\section{PIH with Large Offsets}
\label{app:large_offsets}
\begin{figure*}[h]
    \centering
    \includegraphics[width=\textwidth]{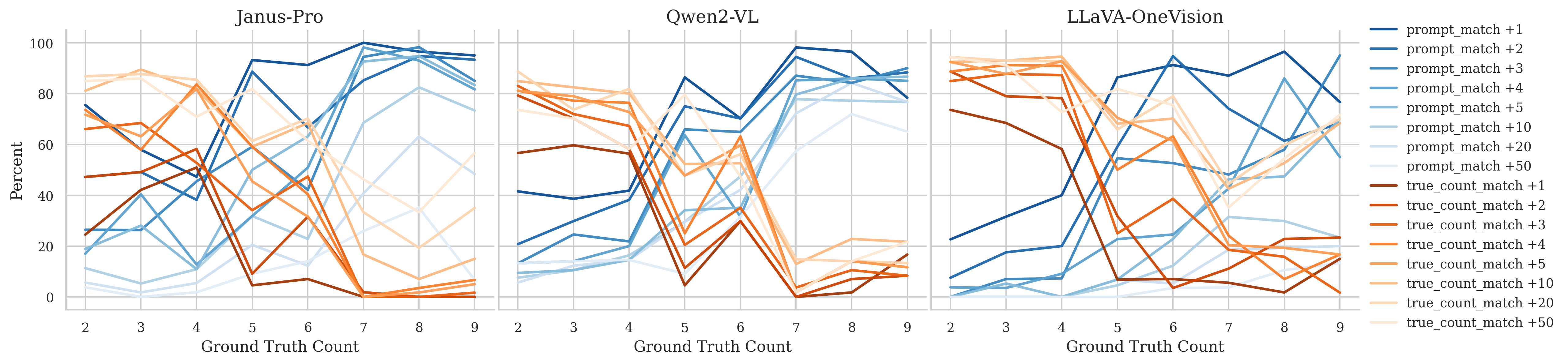}
    \caption{\textbf{PIH rates across different number of ground-truth objects.} Blue lines show the percentage of outputs matching the prompted number for different offsets $k$, while orange lines show matches to the true count. Bias grows with object count, with four objects marking the approximate threshold at which models stop correcting the prompt. Prompt match and true count match do not sum to $100$\% because some model responses do not contain a number.}
    \label{fig:confirmation_50}
\end{figure*}

Figure~\ref{fig:confirmation_50} extends the analysis in Figure~\ref{fig:confirmation} to large discrepancy offsets ($k \in \{10, 20, 50\}$), where the prompted count substantially exceeds the number of objects present in the image. Across all three models, we observe that once the number of ground-truth objects exceeds approximately four, prompt-induced hallucinations persist even under extreme misalignment: models frequently produce responses matching the prompted count, while corrections to the true count become rare.

However, LLaVA-OneVision exhibits a notable deviation from this pattern. While Qwen-VL and Janus-Pro show near complete convergence toward prompt matching at higher object counts, regardless of the offset magnitude, LLaVA-OneVision maintains a wider separation between prompt-aligned and ground-truth-aligned responses. In particular, for offsets of $+10$, $+20$, and $+50$, LLaVA-OneVision continues to produce a non-trivial fraction of corrections to the true count, even when the discrepancy is large.

This behavior indicates that LLaVA-OneVision is more resilient to extreme prompt–image mismatches, retaining some capacity to reject implausible prompt specifications at higher object counts. The persistence of ground-truth corrections under large offsets is consistent with our broader findings that LLaVA-OneVision exhibits stronger visual grounding and reduced reliance on prompt-based copying compared to Qwen-VL and Janus-Pro. Nevertheless, even for LLaVA-OneVision, prompt-induced hallucinations dominate once object counts are sufficiently high, underscoring the strength of prompt-driven biases in this regime.

\section{Base Probabilities for All Models}
\label{app:base_probs_all}
\begin{figure*}
    \centering
    \includegraphics[width=\linewidth]{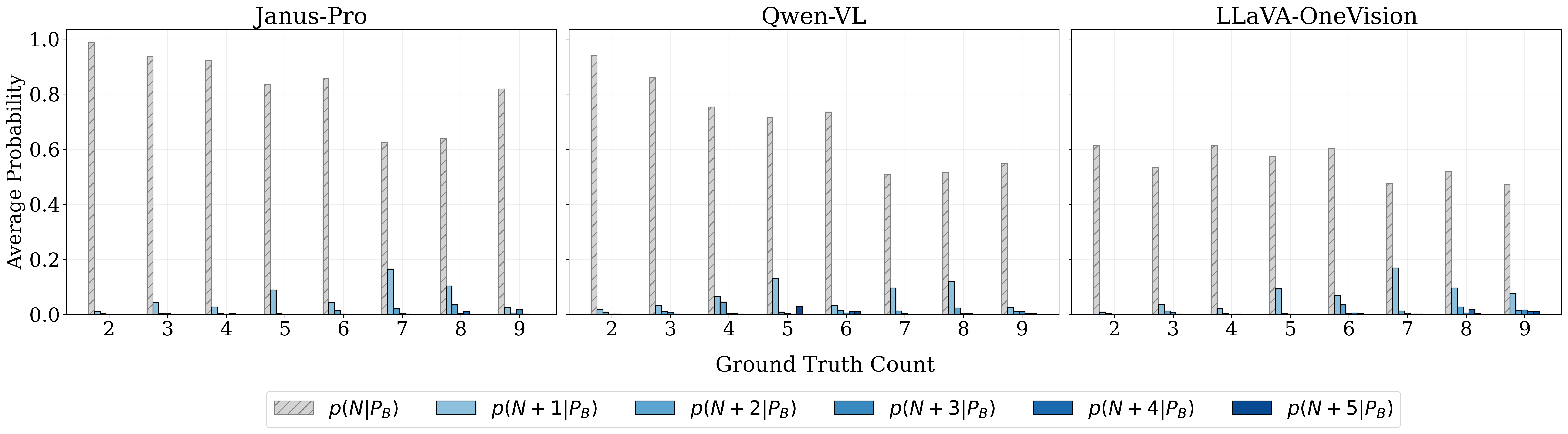}
    \caption{\textbf{Base probabilities on the object counting task.} We plot the probability assigned to the true count $N$ and neighboring counts $N+k$ ($k \in \{1,\ldots,5\}$) given the prompt ``How many [objects] are in the image?''. As object counts increase, confidence in $N$ decreases and probability mass spreads to neighboring counts, matching the increased susceptibility to PIH.}
    \label{fig:app_base_probs}
\end{figure*}

Figure~\ref{fig:app_base_probs} shows the probabilities for the true count $N$ and neighboring counts $N+k$ under the baseline prompt ($P_{B}$) for all models.

\section{Attention Knockout Details}
\label{app:attention_knockout}
To analyze the causal role of individual attention heads in PIH, we perform attention head knockouts using mean ablation \cite{wang2022interpretability, mueller2025mib}.
Let $H^{(l,h)} \in \mathbb{R}^{T \times d}$ denote the output of head $h$ at layer $l$ across $T$ tokens. We compute the head-wise mean
\[
\mu^{(l,h)} = \frac{1}{T} \sum_{t=1}^{T} H^{(l,h)}_t,
\]
and replace the head output at every token position with this mean, i.e.,
\[
\tilde{H}^{(l,h)}_t = \mu^{(l,h)} \quad \forall t \in \{1,\dots,T\}.
\]
This intervention removes token-specific information carried by the head while preserving its overall activation magnitude.

We apply this procedure head-by-head and quantify knockout success as the fraction of samples that switch from the prompted (hallucinated) count to the true image-grounded count under the misaligned prompt.
We rank heads by their corrective effect and select the top-$m$ heads per model.
We then perform a second-stage grouped knockout in which all top-$m$ heads are ablated simultaneously for $m \in \{1,3,5,10\}$, selecting $m$ as a hyperparameter per-model.

\section{Full List of PIH-heads}
\label{app:pih_heads}
\begin{table}[t]
\centering
\small
\begin{tabular}{cccc}
\toprule
\textbf{Model} & \textbf{Rank} & \textbf{Head} & \textbf{Success (\%)} \\
\midrule
\multicolumn{4}{l}{\textbf{Qwen-VL}} \\
 & 1  & L0H3   & 38.90 \\
 & 2  & L0H6   & 26.89 \\
 & 3  & L11H15 & 16.71 \\
 & 4  & L0H11  & 8.09  \\
 & 5  & L15H8  & 7.57  \\
 & 6  & L14H13 & 7.05  \\
 & 7  & L0H10  & 6.79  \\
 & 8  & L7H3   & 6.53  \\
 & 9  & L0H24  & 6.01  \\
 & 10 & L8H13  & 5.48  \\
\midrule
\multicolumn{4}{l}{\textbf{Janus-Pro}} \\
 & 1  & L0H20  & 31.18 \\
 & 2  & L1H7   & 27.96 \\
 & 3  & L14H27 & 17.74 \\
 & 4  & L12H9  & 17.47 \\
 & 5  & L11H18 & 15.59 \\
 & 6  & L13H2  & 13.98 \\
 & 7  & L0H24  & 13.44 \\
 & 8  & L14H28 & 13.17 \\
 & 9  & L8H11  & 12.90 \\
 & 10 & L8H3   & 12.63 \\
\midrule
\multicolumn{4}{l}{\textbf{LLaVA-OneVision}} \\
 & 1  & L0H3   & 48.99 \\
 & 2  & L0H6   & 37.92 \\
 & 3  & L0H26  & 30.20 \\
 & 4  & L11H15 & 27.85 \\
 & 5  & L0H23  & 27.52 \\
 & 6  & L0H24  & 21.48 \\
 & 7  & L0H15  & 20.81 \\
 & 8  & L14H9  & 17.79 \\
 & 9  & L0H11  & 16.78 \\
 & 10 & L17H22 & 15.10 \\
\bottomrule
\end{tabular}
\caption{Top-10 attention heads per model ranked by individual knockout success.
Success is measured as the percentage of samples that switch from PIH to the correct image-grounded count when the head is ablated.}
\label{tab:pih_heads_all_models}
\end{table}

Table~\ref{tab:pih_heads_all_models} reports the top-ranked attention heads for each model, ranked by their individual knockout success. Knockout success is defined as the fraction of misaligned-prompt samples that switch from matching the prompted (hallucinated) count to matching the true image-grounded count when a single attention head is ablated via mean ablation. Higher values therefore indicate a stronger causal role in propagating prompt-induced numerical errors.

For both Janus-Pro and LLaVA-OneVision, performance improves monotonically with larger head sets, with the best results achieved when ablating the top 10 PIH heads. In contrast, Qwen-VL exhibits its highest success rate when ablating only the top 3 heads, with performance degrading substantially when larger groups are removed. This divergence suggests that while PIH in Janus-Pro and LLaVA-OneVision is distributed across a broader set of attention heads, Qwen-VL relies on a smaller subset of highly specialized heads, and removing additional heads begins to interfere with general generation behavior. Accordingly, we select the top-$10$ heads as the final PIH head sets for Janus-Pro and LLaVA-OneVision, and the top-$3$ heads for Qwen-VL in all subsequent experiments.

\begin{table}[t]
\centering
\small
\setlength{\tabcolsep}{6pt}
\renewcommand{\arraystretch}{1.2}
\begin{tabular}{lcccc}
\toprule
\textbf{Model} & \textbf{Best-1} & \textbf{Best-3} & \textbf{Best-5} & \textbf{Best-10} \\
\midrule
Janus-Pro       & 31.18 & 65.06 & 73.19 & 77.91 \\
LLaVA-OV & 48.99 & 72.35 & 73.68 & 83.08 \\
Qwen-VL    & 38.90   & 67.14 & 66.86 & 3.43  \\
\bottomrule
\end{tabular}
\caption{Global mean success rates (\%) under different PIH head selection strategies.
Best-$m$ denotes ablating the top-$,$ PIH heads ranked by individual knockout success.}
\label{tab:global_mean_summary}
\end{table}

\subsection{Attention Patterns of PIH-heads}
\label{app:attention_patterns}

\begin{figure*}[t]
    \centering
    \includegraphics[width=\textwidth]{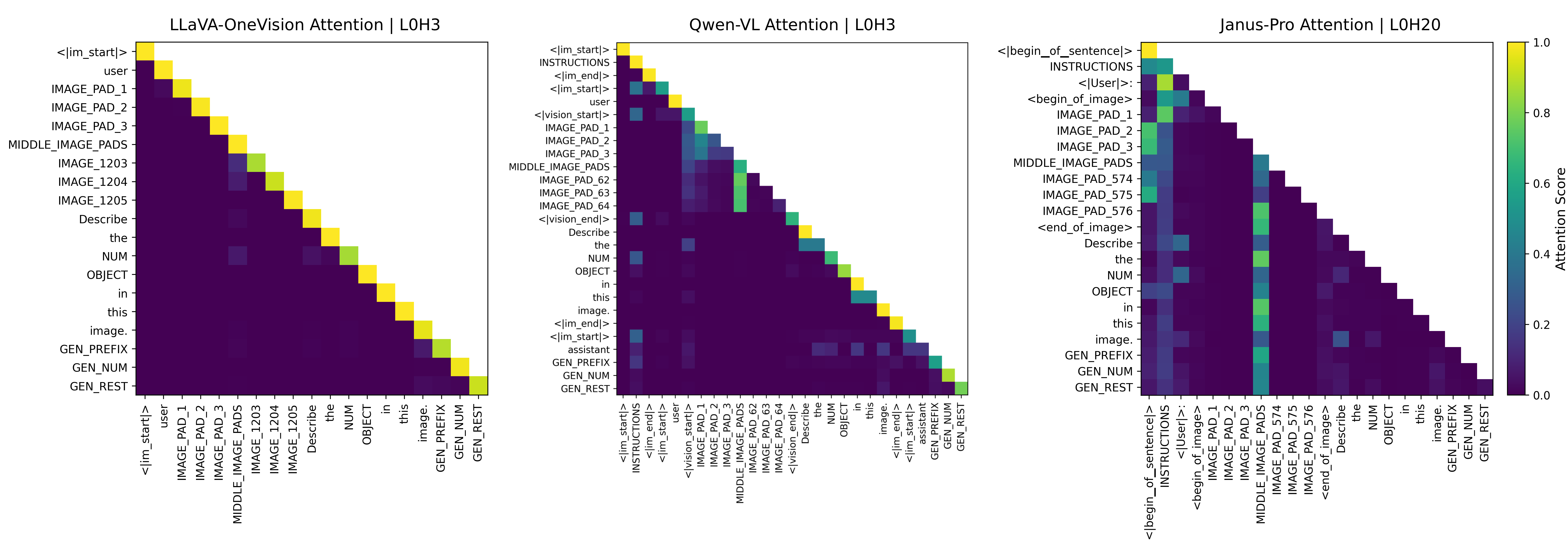}
    \caption{
    Attention patterns for the top-ranked PIH attention head in three multimodal models, averaged over all samples and computed over the full input–generation sequence.
    \textbf{Left:} LLaVA-OneVision (L0H3),
    \textbf{Center:} Qwen-VL (L0H3),
    \textbf{Right:} Janus-Pro (L0H20).
    Query tokens are shown on the y-axis and key tokens on the x-axis, with each row summing to $1$.
    Image pad tokens are largely aggregated into a single row and column (\textit{IMAGE\_PADS}), with boundary tokens retained.
    }
    \label{fig:pih_attn_all_models}
\end{figure*}

To gain qualitative insight into the behavior of PIH attention heads across models, we visualize the attention patterns of the highest-ranked PIH head in each model, Qwen-VL (L0H3), LLaVA-OneVision (L0H3), and Janus-Pro (L0H20), over the full input and generation sequences, averaged across all samples.
Representative examples are shown in \cref{fig:pih_attn_all_models}.

In all visualizations, query tokens are shown on the y-axis and key tokens on the x-axis, such that each row of attention scores sums to $1$. Since images are represented as long sequences of image pad tokens (often several hundred), we aggregate the majority of these tokens into a single row and column, denoted as \textit{IMAGE\_PADS}, by summing their attention scores.
We retain the first three and last three image pads as individual entries to preserve boundary structure and improve interpretability.

The PIH head in Qwen-VL (L0H3) exhibits a mixed attention pattern. A substantial portion of attention lies along the diagonal over the text sequence, indicating behavior similar to local self-attention. At the same time, we observe consistent and structured attention within the image-token region, including attention to the aggregated \textit{IMAGE\_PADS} and nearby vision boundary tokens.
This combination places Qwen-VL L0H3 between purely textual self-attention and fully image-focused attention.

In contrast, the PIH head identified in Janus-Pro (L0H20) shows attention patterns that are predominantly concentrated on image tokens. The diagonal structure characteristic of local textual self-attention is largely absent, and attention is instead focused almost entirely within the image-token region across many query positions.
This behavior suggests that the PIH head in Janus-Pro functions primarily as an image-focused attention head.

The PIH head in LLaVA-OneVision (L0H3) behaves almost entirely as a local self-attention head over text.
Attention is concentrated along the diagonal, with little to no direct attention to image tokens in the unablated model. Ablation of the PIH head in LLaVA-OneVision leads to the strongest performance improvements among the three models. One possible explanation is that the targeted PIH head in LLaVA-OneVision is not primarily focused on image tokens, but instead behaves as a largely textual self-attention head. As a result, ablating this head may predominantly disrupt copying or local textual interactions, without directly impairing image processing.
More generally, ablation may induce second-order effects across all models, whereby other heads adjust their attention patterns and potentially increase attention to image tokens. In the case of LLaVA-OneVision, we hypothesize that the combination of not removing an image-focused head and benefiting from such second-order redistribution effects contributes to its robustness and improved performance under PIH ablation.

For all three models, we do not observe a clear, isolated peak of attention on the generated number token, so while some of the heatmap are suggestive of modality-specific routing, none are conclusive about how the numeric value is produced. This inconclusiveness is reasonable, as interpreting the internal function of attention heads from visualization alone remains a well-known challenge despite their demonstrated causal effects \cite{jain2019attention, serrano2019attention, brunner2019identifiability, grimsley2020attention}.

\section{Output Examples After Knockouts}
\label{app:generation_examples}

Table \ref{tab:app_qualitative_knockout_examples} shows examples of generated texts after PIH-head ablations.

\begin{table*}[t]
\centering
\scriptsize
\setlength{\tabcolsep}{6pt}
\renewcommand{\arraystretch}{1.15}
\begin{tabular}{p{7.2cm} p{7.2cm}}
\toprule
\textbf{Before PIH Knockout} & \textbf{After PIH Knockout} \\
\midrule

\multicolumn{2}{l}{\textbf{Qwen-VL}} \\[0.3em]

\begin{itemize}[leftmargin=*, itemsep=0.6em, topsep=0pt]
\item The image shows a set of \textbf{52} glasses, which appear to be stemless wine glasses. They are filled with red wine, and the glasses are arranged in a grid pattern, with \textbf{13 rows and 4 columns}.
\item The image shows \textbf{three} pairs of leggings. The first pair is light pink, the second pair is white with a small pattern, and the \textbf{third pair} is also white but without any pattern.
\item The image shows a total of \textbf{four} wineglasses. \textbf{Two} of them are filled with red wine and are placed side by side on a wooden table. \textbf{The other two} wineglasses are empty and are also placed on the table.
\end{itemize}
&
\begin{itemize}[leftmargin=*, itemsep=0.6em, topsep=0pt]
\item The image shows \textbf{two} wine glasses. The glasses are stemless and have a modern, minimalist design. They are filled with red wine, and the glasses appear to be made of clear glass.
\item The image shows \textbf{two} pairs of leggings. The pair on the left is light pink, and the pair on the right is white with a subtle pattern.
\item The image shows \textbf{two} wine glasses filled with red wine. The glasses are placed on a wooden table, and there is a bottle of red wine and a basket of grapes in the background.
\end{itemize}
\\[0.8em]

\midrule
\multicolumn{2}{l}{\textbf{Janus-Pro}} \\[0.3em]

\begin{itemize}[leftmargin=*, itemsep=0.6em, topsep=0pt]
\item The image shows \textbf{four} kids standing together outdoors. They are all wearing casual clothing, with \textbf{one} of them in a plaid shirt over a yellow t-shirt. \textbf{The other three} are in solid-colored shirts.
\item The image shows \textbf{four} croissants, which are golden brown in color and have a flaky, layered texture. They are arranged on a wooden cutting board, with \textbf{two} croissants placed side by side and the \textbf{other two} slightly overlapping each other.
\item The image shows \textbf{9} planes flying in formation. They are all blue with white markings and have the word ``Canada'' written on them. The planes are arranged in a tight, symmetrical pattern, with \textbf{four} planes on the top row and \textbf{five} on the bottom.
\end{itemize}
&
\begin{itemize}[leftmargin=*, itemsep=0.6em, topsep=0pt]
\item The image shows \textbf{two} children standing close together outdoors. The child on the left has curly hair and is wearing a bright orange shirt with a red and blue checkered shirt over it. The child on the right has short hair and is wearing a dark shirt.
\item The image shows \textbf{two} croissants placed on a wooden cutting board. They are golden brown, flaky, and have a slightly twisted shape, typical of traditional croissants.
\item The image shows \textbf{four} blue aerobatic planes flying in formation. The planes are marked with the letters ``AC'' and have a distinctive design with white accents.
\end{itemize}
\\[0.8em]

\midrule
\multicolumn{2}{l}{\textbf{LLaVA-OneVision}} \\[0.3em]

\begin{itemize}[leftmargin=*, itemsep=0.6em, topsep=0pt]
\item The image displays \textbf{five} different illustrations of classic convertible cars arranged in a \textbf{two-by-three} grid. Each car is depicted in a different color and style, showcasing a variety of vintage designs.
\item The image shows a set of \textbf{eight} forks arranged in \textbf{two rows} against a blue background. Each fork has a polished, reflective surface, indicating they are made of a shiny metal, likely silver or a silver-plated material.
\item The image displays a collection of \textbf{17} plastic water bottles arranged in a row against a white background. Each bottle appears to be filled with water, and they vary in shape and size.
\end{itemize}
&
\begin{itemize}[leftmargin=*, itemsep=0.6em, topsep=0pt]
\item The image presents a collection of \textbf{four} vintage cars, each distinct in color and model, arranged in a \textbf{two-by-two} grid.
\item The image shows a set of \textbf{six} ornate silver forks arranged in \textbf{a row} against a blue background. Each fork has a polished finish and features intricate designs on the handles.
\item The image displays a collection of \textbf{seven} plastic water bottles arranged in a row against a white background, with cap colors \textbf{pink, white, blue, white, blue, white, and white}.
\end{itemize}
\\

\bottomrule
\end{tabular}
\caption{Qualitative examples of generations before and after PIH attention head knockout.
Before intervention, models not only copy the incorrect count implied by the misaligned prompt, but also hallucinate non-existing details.
After ablating PIH heads, generations more consistently reflect the true number of objects present in the image across all three models.}
\label{tab:app_qualitative_knockout_examples}
\end{table*}

\section{Color Task}
\label{app:colors}
To investigate whether PIH heads generalize beyond the counting task, we extend our framework to a color prediction task using the Visual CounterFact dataset \cite{pvp}.
This setting replaces numerical offsets with color offsets while preserving the overall experimental structure used in counting.
We use ''What color is the [object] in the image?'' as the baseline prompt, and ``Describe the $C+k$ [object]'' as the misaligned prompt, where $C$ denotes the ground-truth color of the object in the image and $k \in \{1,2,3\}$ denotes the perceptual ``difference'' from the $C$ on the color wheel.
In particular, $k=1$ indicates a highly similar color and $k=3$ indicates the \textit{contrasting color} which is the further distance on the color wheel from $C$. 
For example, given $C= \text{red}$, we have orange, yellow, and green for $k=1,2,3$, respectively.

\begin{figure*}
    \centering
    \includegraphics[width=\textwidth]{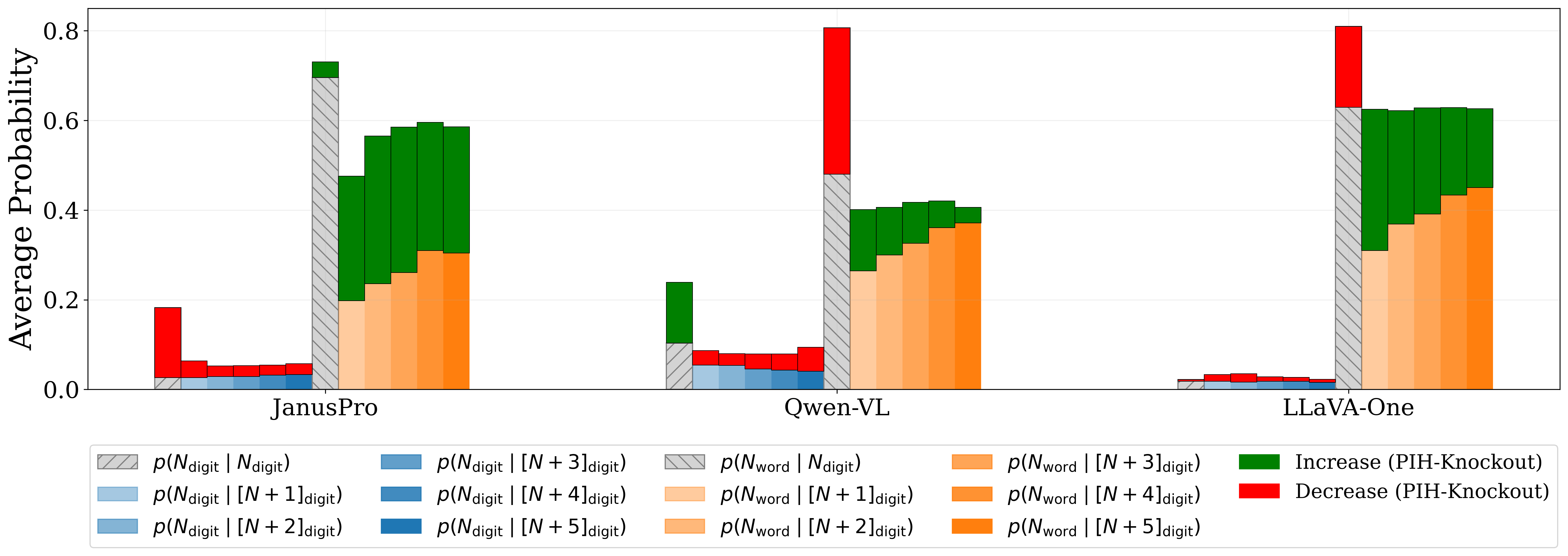}
    \caption{
    Impact of knockout on the probability of the ground-truth answer in both digit (blue) and word form (orange) given the prompt ``Describe the $[N + k]_{\text{digit}}$ [objects]'' for count. Note that when $k=0$, there is no modality conflict in the prompt.
    }
    \label{fig:app_knockout_prob_impact}
\end{figure*}

\section{Layer-wise Attention Mass Shifts After PIH Ablation}
\label{app:attention_mass}

\begin{figure*}[t]
    \centering
    \includegraphics[width=\linewidth]{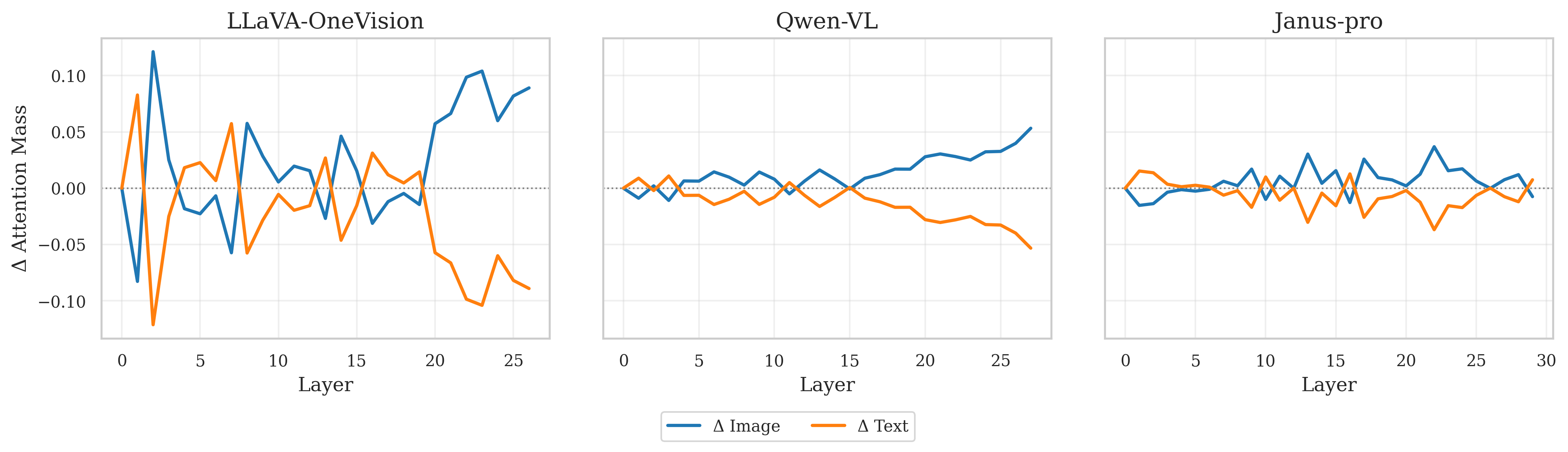}
    \caption{
        \textbf{Layer-wise change in attention mass after PIH head ablation.}
        For each model, we plot the change in attention mass to image tokens (blue) and text tokens (orange) as a function of transformer layer.
        Positive values indicate increased attention allocation relative to the baseline.
        Across all models, PIH ablation consistently shifts attention from text to image tokens, with LLaVA-OneVision exhibiting the strongest and earliest effect.
    }
    \label{fig:delta_attention_per_layer}
\end{figure*}

While \cref{fig:max_delta_layer_image_attention} focuses on the single layer exhibiting the largest intervention-induced shift in attention mass, we additionally examine how PIH head ablation affects attention allocation \emph{across all transformer layers}. 
Specifically, for each layer we compute the change in attention mass to image tokens relative to the baseline, with the complementary change applied to text tokens (since attention mass sums to one).

\cref{fig:delta_attention_per_layer} shows the layer-wise change in image and text attention mass for LLaVA-OneVision, Qwen-VL, and Janus-pro. Across all models, PIH ablation consistently increases attention to image tokens while reducing attention to text tokens, indicating a systematic rebalancing of multimodal attention rather than a highly localized effect. 

Notably, the magnitude and depth of this shift varies across architectures. LLaVA-OneVision exhibits a pronounced increase in image attention in early layers, whereas Qwen-VL and Janus-pro show smaller but more distributed changes concentrated in later layers. These trends suggest that PIH head ablation alters the global flow of multimodal information throughout the network.

\end{document}